\documentclass[11pt]{article}

\PassOptionsToPackage{table}{xcolor}
\usepackage[final]{acl}

\usepackage{times}
\usepackage{latexsym}

\usepackage[T1]{fontenc}
\usepackage[utf8]{inputenc}

\usepackage{microtype}

\usepackage{graphicx}

\usepackage{amsmath}
\usepackage{amssymb}
\usepackage{subcaption}    % 子图
\usepackage{booktabs}      % 专业表格线
\usepackage[table]{xcolor} % 彩色表格行
\usepackage{multirow}      % 表格单元格合并
\usepackage{listings}      % 代码块
\usepackage{mathtools}     % 高级数学公式
\usepackage{amsthm}        % 定理环境
\usepackage{hyperref}      % 超链接
\usepackage[capitalize,noabbrev]{cleveref} % 智能引用
\usepackage[textsize=tiny]{todonotes}      % todo注释
\usepackage{algorithm}
\usepackage{algorithmic}
\usepackage{makecell}
\usepackage{caption}
\title{V-Retrver: Evidence-Driven Agentic Reasoning for Universal Multimodal Retrieval}

\newcommand{\homepage}{\raisebox{-1.5pt}{\includegraphics[height=1em]{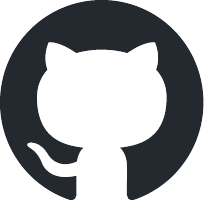}}}
\newcommand{\hfmodel}{\raisebox{-1.5pt}{\includegraphics[height=1em]{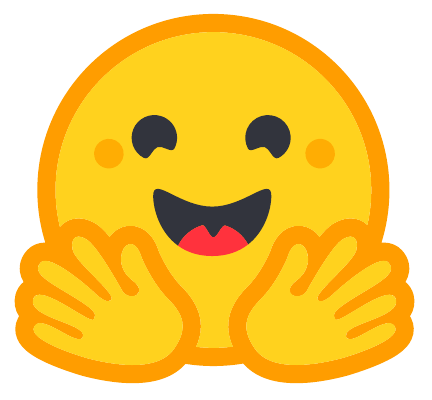}}}

\author{
\bf{
Dongyang Chen$^{1}$\thanks{Equal contribution.},
Chaoyang Wang$^{2\ast}$, Dezhao Su$^{3}$, Xi Xiao$^{1}$\thanks{Corresponding author.}, Zeyu Zhang$^{4}$,}\\
\bf{
Jing Xiong$^{5}$, Qing Li$^{6}$, Yuzhang Shang$^{2}$, Shichao Kan$^{7}$}\\[2mm]
{$^1$Tsinghua University}\ \ 
{$^2$University of Central Florida}\ \ 
{$^3$Fudan University}\\
{$^4$The Australian National University}\ \ 
{$^5$The University of Hong Kong}\\
{$^6$Pengcheng Laboratory}\ \ 
{$^7$Central South University}\\[2mm]
{\homepage\ \normalfont\texttt{Home: }\url{https://github.com/chendy25/V-Retrver}}\\[2pt]
{\hfmodel\ \normalfont\texttt{HF: }\url{https://huggingface.co/V-Retrver}}
}

\begin{document}
% Columns are not force-stretched: slack goes to the column bottom instead of
% inflating the gap before a \paragraph heading.
\raggedbottom
\maketitle 

% \begin{abstract}
% Multimodal Large Language Models (MLLMs) have recently been applied to universal multimodal retrieval, where Chain-of-Thought (CoT) reasoning improves candidate reranking. However, existing approaches remain largely language-driven, relying on static visual encodings and lacking the ability to actively verify fine-grained visual evidence, which often leads to speculative reasoning in visually ambiguous cases. We propose V-Retrver, an evidence-driven retrieval framework that reformulates multimodal retrieval as an agentic reasoning process grounded in visual inspection. V-Retrver enables an MLLM to selectively acquire visual evidence during reasoning via external visual tools, performing a multimodal interleaved reasoning process that alternates between hypothesis generation and targeted visual verification. To train such an evidence-gathering retrieval agent, we adopt a curriculum-based learning strategy combining supervised reasoning activation, rejection-based refinement, and reinforcement learning with an evidence-aligned objective. Experiments across multiple multimodal retrieval benchmarks demonstrate consistent improvements in retrieval accuracy (with 23.0\% improvements on average), perception-driven reasoning reliability, and generalization.
% \end{abstract}

\begin{abstract}
% Multimodal Large Language Models (MLLMs) have recently been adopted for universal
% multimodal retrieval, showing that Chain-of-Thought (CoT) reasoning can improve candidate selection. However, existing CoT-based retrieval frameworks remain predominantly language-driven, relying on static visual encodings and thus failing to utilize fine-grained visual evidence. This often leads to speculative or hallucinated reasoning, especially when candidate images share similar semantics but differ in subtle visual attributes.
% To address this limitation, we propose V-Retrver, the first O3-like MLLM for universal multimodal retrieval. V-Retrver empowers an MLLM to actively gather visual evidence during reasoning by invoking external visual tools. Through an  multimodal interleaved CoT process, the model dynamically inspects candidate images, verifies visual hypotheses, and progressively refines ranking decisions. 
% To incentivize  this capability, we design a three-stage training framework: (1) a cold-start training stage using synthesized data to establish basic reasoning capabilities; (2) Rejection Sampling Fine-Tuning to further enhance the model’s reasoning ability; and (3) Group Relative Policy Optimization (GRPO) to continuously reinforce reasoning performance.
% Experimental results demonstrate that V-Retrver achieves substantial performance improvements across multiple benchmarks and tasks, exhibiting stronger perception-driven reasoning, higher retrieval accuracy, and better generalization ability.
Multimodal Large Language Models (MLLMs) have recently been applied to universal multimodal retrieval, where Chain-of-Thought (CoT) reasoning improves candidate reranking. However, existing approaches remain largely language-driven, relying on static visual encodings and lacking the ability to actively verify fine-grained visual evidence, which often leads to speculative reasoning in visually ambiguous cases. We propose V-Retrver, an evidence-driven retrieval framework that reformulates multimodal retrieval as an agentic reasoning process grounded in visual inspection. V-Retrver enables an MLLM to selectively acquire visual evidence during reasoning via external visual tools, performing a multimodal interleaved reasoning process that alternates between hypothesis generation and targeted visual verification. To train such an evidence-gathering retrieval agent, we adopt a curriculum-based learning strategy combining supervised reasoning activation, rejection-based refinement, and reinforcement learning with an evidence-aligned objective. Experiments across multiple multimodal retrieval benchmarks demonstrate consistent improvements in retrieval accuracy (with 23.0\% improvements on average), perception-driven reasoning reliability, and generalization.
\end{abstract}

% \weidi{not impressive enough, give numbers, how many different retrieval tasks, datasets, blabla}
\section{Introduction}

\begin{figure*}[t!]
    \centering
    \includegraphics[width=0.99\linewidth]{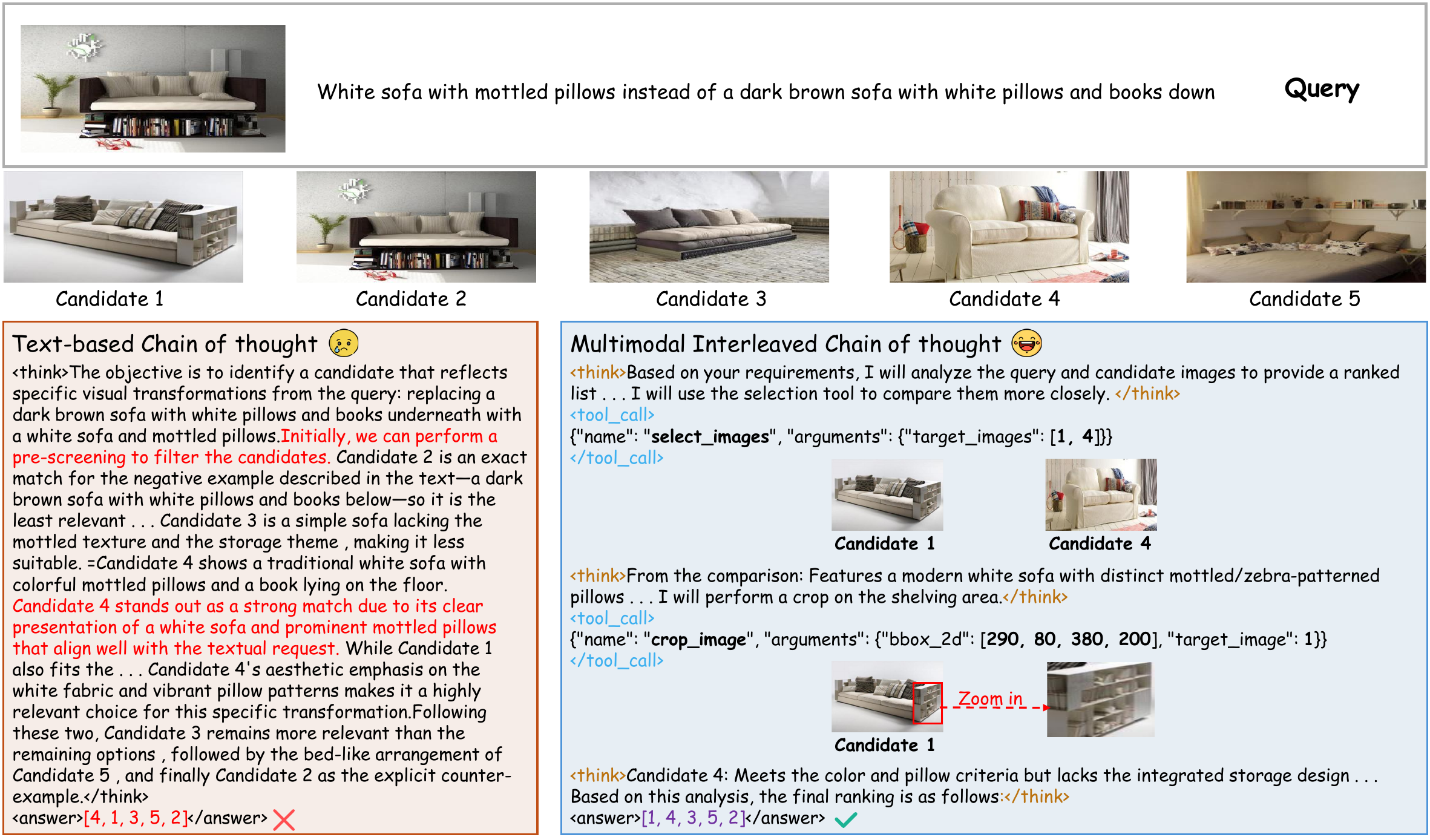}
    \caption{Comparison between text-based CoT (left) and multimodal interleaved CoT (right) for multimodal retrieval. Text-based CoT relies on language-driven inference over static visual representations, often failing to resolve fine-grained differences. In contrast, V-Retrver performs multimodal interleaved CoT reasoning by invoking visual tools to inspect candidate images, enabling grounded reasoning and more reliable ranking decisions.}
    \label{fig:intro}
    \vspace{-0.45cm}
\end{figure*}

The rapid development of Multimodal Large Language Models (MLLMs) has substantially advanced universal multimodal retrieval~\cite{chen2024mllm, lin2024mm, wang2024multimodal, zhu2025retrv}, enabling a single model to support diverse retrieval scenarios such as text-to-image, image-to-text, and interleaved multimodal queries. Recent works further demonstrate that incorporating Chain-of-Thought (CoT) reasoning can improve retrieval performance by enhancing interpretability and candidate discrimination~\cite{zhu2025retrv, Xu2025MMR5, narayan2025deepmmsearch}. However, despite these advances, existing CoT-based retrieval systems remain fundamentally language-driven, even when retrieval decisions critically depend on visual evidence.

This limitation becomes particularly pronounced in visually ambiguous retrieval scenarios, where candidate images share similar semantic content but differ in fine-grained visual attributes such as object appearance, style, or local context. Most current MLLM-based retrieval methods \cite{liu2024lamra, chen2024mllm, lin2024mm} compress visual inputs into fixed embeddings or textual descriptions, forcing the reasoning process to rely on language alone to infer visual differences. Consequently, the model often produces speculative or hallucinated reasoning when the required evidence lies in the visual modality. Even recent reasoning-enhanced retrieval frameworks, such as Retrv-R1~\cite{zhu2025retrv} and MM-R5 \cite{Xu2025MMR5}, improve textual reasoning depth but still rely on single-pass visual encoding, lacking the ability to actively verify visual hypotheses during reasoning.

To overcome this gap, we propose \textbf{V-Retrver}, an evidence-driven retrieval framework that reformulates multimodal retrieval as an agentic reasoning process grounded in visual inspection. Instead of treating visual representations as static inputs, V-Retrver enables an MLLM to selectively acquire visual evidence during reasoning by invoking external visual tools. Through a multimodal interleaved Chain-of-Thought process, the model alternates between hypothesis generation and targeted visual verification, allowing it to dynamically resolve visual ambiguities and progressively refine ranking decisions, as illustrated in Fig.~\ref{fig:intro}.

Training such an evidence-gathering retrieval agent requires not only strong reasoning ability but also effective alignment between retrieval performance and visual tool usage. We therefore adopt a curriculum-based training strategy consisting of three stages. First, a cold-start supervised stage initializes the model with basic reasoning capabilities and operation formatting using synthesized high-quality CoT data. Second, rejection sampling fine-tuning consolidates high-quality reasoning trajectories and improves structural compliance. Finally, we introduce \textbf{Evidence-Aligned Policy Optimization (EAPO)}, instantiated via Group Relative Policy Optimization (GRPO)~\cite{guo2025deepseek}, which reinforces correct ranking decisions while encouraging informative visual verification and discouraging redundant tool usage.

Extensive experiments on the universal multimodal retrieval benchmark M-BEIR, as well as multiple out-of-domain datasets, demonstrate that V-Retrver consistently outperforms strong baselines across diverse retrieval settings. The results show that V-Retrver achieves higher retrieval accuracy, more reliable perception-grounded reasoning, and stronger generalization ability, validating the effectiveness of interleaved visual reasoning for multimodal retrieval. In summary, our contributions are three-fold:
\begin{itemize}
    \item We propose V-Retrver, an evidence-driven agentic retrieval framework that enables MLLMs to actively acquire visual evidence during multimodal reasoning.
    \item We introduce a curriculum-based training strategy with an evidence-aligned reinforcement learning objective that jointly improves reasoning quality, ranking accuracy, and efficient visual tool usage.
    \item Extensive experiments across multiple benchmarks demonstrate that V-Retrver consistently outperforms existing methods and generalizes well to diverse multimodal retrieval scenarios.
\end{itemize}

\section{Related Work}

\paragraph{Multi-modal Large Language Models.}
In recent years, the rapid advancement of MLLMs has driven the deep integration of visual perception and language reasoning, leading to the emergence of high-performing open-source models, notably the LLaVA~\cite{liu2024llava,guo2024llava,zhang2025llava,lin2023video,li2023llava}, Qwen-VL~\cite{bai2023qwenvl,wang2024qwen2vl,bai2025qwen2}, and InternVL~\cite{chen2024internvl,gao2024mini,lu2025internvl} series. In parallel, large-scale models such as Flamingo~\cite{alayrac2022flamingo}, mPLUG-Owl~\cite{ye2023mplugowl,ye2024mplugowl2,ye2024mplug3}, and GPT-4V~\cite{yang2023dawn} pursue a holistic vision-language modeling paradigm, incorporating advanced mechanisms including mixture-of-experts~\cite{shu2024llava,li2025uni} and image generation~\cite{xie2024show,xu2025show}. However, these models generally lack Chain-of-Thought reasoning and test-time scalability~\cite{muennighoff2025s1,zhang2025and,chen2024expanding}, and still largely decouple visual perception from language reasoning.

\vspace{2pt}
\paragraph{Multimodal Retrieval.}
Recent advances in deep learning~\cite{zhu2021learning,zhu2024llafs,zhu2025llafs++,zhu2025replay,zhu2025not,ji2024discrete} have driven progress across retrieval tasks, including text–image retrieval~\cite{pham2024composing,fu2024linguistic,zhang2020context,chun2021probabilistic,kim2023exposing,kim2023improving}, composed image retrieval~\cite{baldrati2022effective,saito2023pic2word,gu2024language,suo2024knowledge,baldrati2023zero}, multimodal document retrieval~\cite{chen2023can,hu2023open,liuuniversal}, and instruction-based retrieval~\cite{wu2021fashion,zhang2024magiclens,asai2023task}. Among these, CLIP~\cite{radford2021learning} has demonstrated strong effectiveness and scalability in multimodal retrieval~\cite{baldrati2022effective,wei2024uniir,sain2023clip,pei2023clipping,jin2024end}. For instance, \citet{kim2023improving} represent each sample as a set of diverse embeddings to better handle cross-modal ambiguity.
More recently, MLLMs have been introduced to advance retrieval~\cite{liu2024lamra,jiang2024e5,lin2024mm,zhou2024megapairs}: some use MLLM embeddings for similarity-based retrieval~\cite{zhou2024megapairs,lan2025llave,lin2024mm,zhang2024gme,jian2025rzenembed,gu2025breaking}, while others employ MLLMs as reranking agents~\cite{liu2024lamra,li2025u}. Retrv-R1~\cite{zhu2025retrv} further equips the model with text reasoning via reinforcement learning.
In contrast, V-Retrver introduces an evidence-driven framework that adaptively invokes visual tools during reasoning, enabling a more flexible and effective process with significant retrieval improvements.

\section{Method}

\begin{figure*}[t!]
    \centering
    \includegraphics[width=\linewidth]{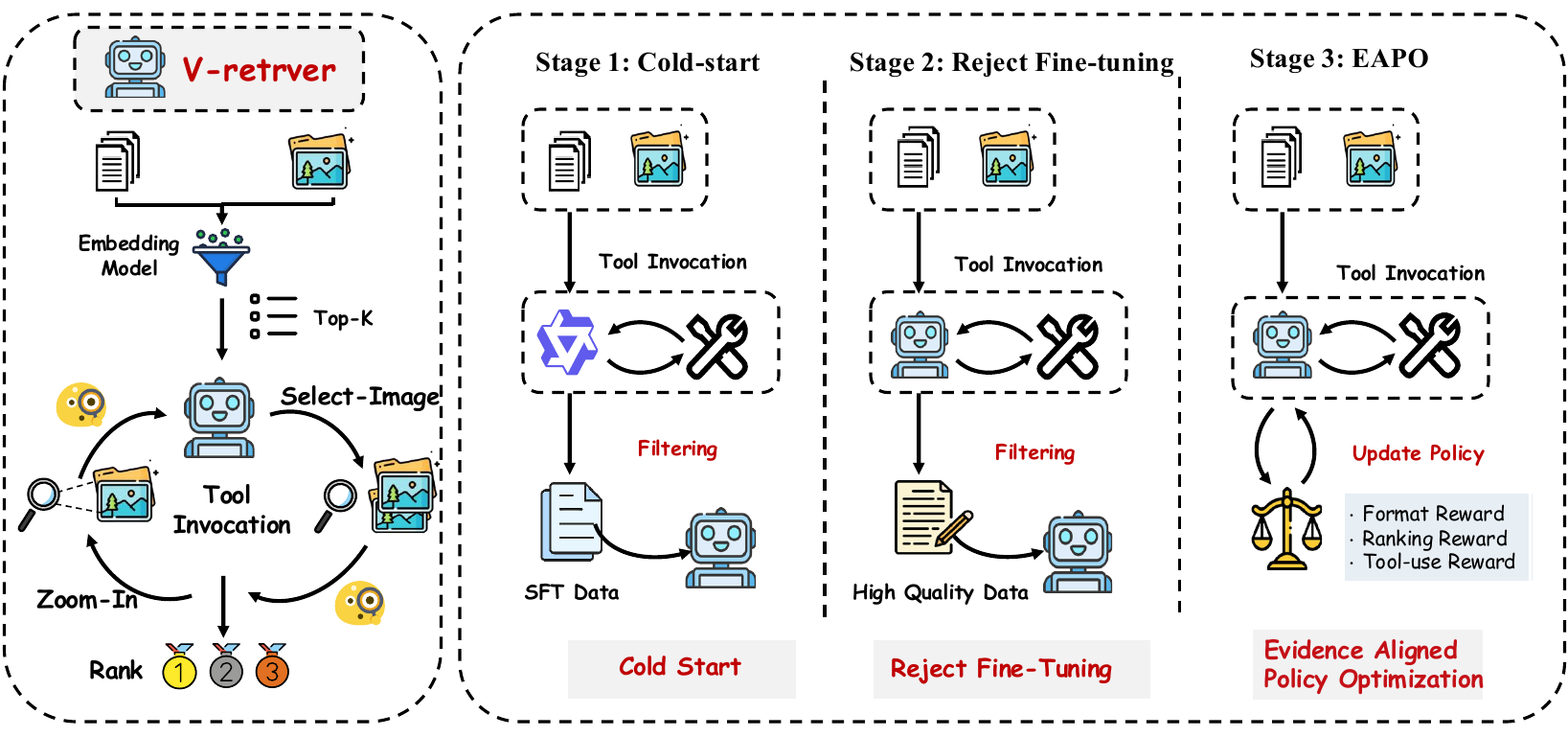}
\caption{Overview of the V-Retrver framework. The left panel illustrates the inference pipeline, featuring a coarse-to-fine process with embedding-based retrieval and agentic reranking. The right panel details the three training stages we proposed, including Cold Start, Rejection sampling Fine-Tuning, and EAPO.}
    \label{fig:method}
    \vspace{-4mm}
\end{figure*}

\subsection{Problem Formulation}
\label{text_overview}
We study the problem of \emph{universal multimodal retrieval}. Given a query $q$ of arbitrary modality (text, image, or interleaved multimodal input) and a candidate pool $\Omega=\{c_n\}_{n=1}^N$, the objective is to identify the most relevant candidate $\hat{c}\in\Omega$. Conventional multimodal retrieval approaches typically formulate this problem as static similarity matching or language-only reranking over fixed visual representations. Such formulations implicitly assume that all necessary visual evidence has been fully encoded into embeddings or textual descriptions \emph{prior} to reasoning. However, this assumption breaks down in fine-grained or visually ambiguous retrieval scenarios, where subtle local details determine relevance and cannot be reliably inferred from compressed representations alone.

To address this limitation, we reformulate multimodal retrieval as an \emph{evidence-grounded reasoning problem}. Under this formulation, retrieval is no longer a single-pass inference process, but an iterative decision-making procedure in which the model is required to actively acquire and verify visual evidence during ranking. Specifically, the retrieval process consists of three tightly coupled steps:
(i) generating hypotheses about candidate relevance based on available information,
(ii) selectively inspecting visual evidence to resolve uncertainty, and
(iii) refining the ranking decision based on verified observations.
This perspective naturally gives rise to an \emph{agentic reranking} paradigm, where a retrieval model is endowed with the ability to reason, inspect, and revise its decisions, rather than passively scoring candidates using fixed representations.

\subsection{Overview of V-Retrver}

Building on the above formulation, we propose \textbf{V-Retrver}, an evidence-driven reasoning framework for universal multimodal retrieval, As illustrated in Fig. \ref{fig:method}. V-Retrver follows a coarse-to-fine retrieval pipeline that decouples efficient candidate proposal from computationally intensive evidence-based reasoning. In the first stage, an embedding model $\phi$ encodes the query $q$ and each candidate $c_n$ into a shared representation space, retrieving the top-$K$ candidates based on similarity. We adopt the same method as LamRA ~\cite{liu2024lamra} for constructing the embedding model $\phi$. This stage serves as an efficient candidate proposal mechanism and substantially reduces the search space:
\[
\mathcal{C} = \{c_k\}_{k=1}^{K}, \quad K \ll N.
\]
In the second stage, V-Retrver employs a reasoning agent $\theta$ to perform fine-grained reranking over the reduced candidate set $\mathcal{C}$. Crucially, $\theta$ is not a conventional reranker that operates over static features. Instead, it is designed as an \emph{agentic evidence-gathering model} that can iteratively reason, invoke visual inspection tools, and revise its ranking decisions based on newly acquired visual observations. The final prediction is produced as:
\[
\hat{c} = \theta(q, \mathcal{C}).
\]

The remainder of this section details the core mechanisms that enable evidence-driven reasoning in V-Retrver, including multimodal interleaved reasoning, visual tools, and a curriculum-based training strategy.

\subsection{Multimodal Interleaved Evidence Reasoning}
\label{sec:mier}

We introduce \textbf{Multimodal Interleaved Evidence Reasoning (MIER)}, a reasoning paradigm that tightly interleaves textual hypothesis generation with targeted visual evidence acquisition. Unlike language-only Chain-of-Thought reasoning, MIER allows intermediate reasoning steps to be explicitly grounded in visual observations obtained on demand.
Formally, given an initial textual query $T_0$ and a candidate image set $I_0$, the reasoning agent iteratively produces outputs:
\[
O_k = f_{\text{MLLM}}\big(\{T_i, C_i, V_i\}_{i=0}^{k}\big),
\]
where $T_i$ denotes a textual reasoning step, $C_i$ denotes a tool invocation request, and $V_i$ represents the visual evidence returned by the tool. A parser then determines whether to extract the next reasoning step and tool request $(T_{k+1}, C_{k+1})$, or to terminate the process and output a final ranking.

If a tool is invoked, the corresponding visual tool is executed and returns new visual evidence $V_{k+1}$, which is appended to the reasoning context. This process yields a multimodal reasoning trajectory:
\[
\tau = \{T_1, C_1, V_1, T_2, C_2, V_2, \dots, T_n, A_n\},
\]
where $A_n$ denotes the final ranked list of candidates.
By explicitly grounding intermediate reasoning steps in dynamically acquired visual evidence, MIER mitigates speculative inference and hallucination, enabling more reliable ranking decisions in visually ambiguous cases.

\subsection{Visual Tools}
To support MIER, we equip the reasoning agent with a set of \textbf{Visual Tools}, which serve as external perceptual interfaces for selective visual inspection. These tools allow the model to control \emph{what} to observe and \emph{where} to focus during reasoning.
Specifically, we implement two tools:

 \textbf{(1) SELECT-IMAGE}, which enables the agent to select a subset of candidate images for closer inspection when multiple candidates exhibit high semantic similarity.
 
 \textbf{(2) ZOOM-IN}, which performs localized zoom-in operations on specified regions of an image, allowing fine-grained analysis of discriminative visual attributes such as objects, textures, or spatial configurations.

These tools facilitate \emph{selective perception} during retrieval. Rather than encoding all visual information upfront, the agent dynamically expands its visual receptive field only when necessary, closely mirroring human retrieval behavior in which ambiguous candidates are resolved by ``looking again'' at critical details.

\subsection{Training V-Retrver via Curriculum-Based Agentic Learning}

Training V-Retrver requires transforming a general-purpose MLLM into an agent capable of stable, evidence-driven reasoning and strategic tool usage. To this end, we design a three-stage curriculum that progressively builds reasoning structure, reliability, and decision-making optimality.

\textbf{Stage I: Reasoning Activation via Supervised Fine-Tuning.}
We begin with a cold-start supervised fine-tuning stage to activate basic reasoning and tool-use behaviors. Since existing retrieval datasets lack annotated reasoning trajectories, we synthesize multimodal Chain-of-Thought data using Qwen2.5-VL-72B-Instruct. These trajectories include structured reasoning steps and valid tool invocation patterns. After applying rule-based filtering to remove logically inconsistent or malformed samples, the base model is fine-tuned using standard SFT loss. This stage establishes foundational reasoning syntax and tool awareness, but does not yet guarantee robustness or optimal tool-use strategies.

\textbf{Stage II: Rejection Fine-Tuning for Reasoning Reliability.}
Although Stage I activates tool-use behavior, the resulting policy exhibits high variance and produces a large fraction of low-quality trajectories. To improve reasoning reliability, we perform Rejection Sampling Fine-Tuning (RSFT). For each training instance, we sample multiple reasoning trajectories and retain only those that strictly satisfy formatting constraints and yield correct retrieval rankings. Fine-tuning on this filtered dataset significantly improves logical consistency and format compliance, providing a stable initialization for reinforcement learning.

\textbf{Stage III: Evidence-Aligned Policy Optimization.}
While the previous stages activate structured reasoning and improve trajectory reliability, they do not explicitly optimize \emph{how} visual evidence should be acquired during retrieval. In practice, the model may either underutilize visual inspection or invoke tools redundantly without contributing to better ranking decisions. To address this limitation, we introduce \textbf{Evidence-Aligned Policy Optimization (EAPO)}, a reinforcement learning objective that explicitly aligns retrieval performance with effective and economical visual verification behavior.

EAPO formulates multimodal retrieval as a trajectory-level decision-making problem, where each reasoning trajectory $o_i$ is evaluated based on both ranking quality and evidence utilization. Specifically, we define a composite reward:
\begin{equation}
R_i
=
\alpha r_{\text{format}}(o_i)
+
\beta r_{\text{rank}}(o_i)
+
r_{\text{tool}}(o_i),
\end{equation}
where the three components respectively encourage structural correctness, accurate ranking, and informative visual inspection. Below, we detail each reward term.

\emph{Format Compliance Reward.}
The format compliance reward $r_{\text{format}}$ ensures that the model adheres to the required reasoning and output protocols, which is essential for stable policy optimization with structured multimodal outputs.
Let $\Omega_{\text{tag}}$ denote the set of trajectories whose outputs are correctly enclosed by predefined \texttt{<think>} and \texttt{<answer>} tags, and let $\Omega_{\text{list}}$ denote the set of trajectories whose final answers strictly follow the required integer ranking list format. We define:
\begin{equation}
r_{\text{format}}(o_i)
=
\frac{1}{2}\,\mathbb{I}_{\{o_i \in \Omega_{\text{tag}}\}}
+
\frac{1}{2}\,\mathbb{I}_{\{o_i \in \Omega_{\text{list}}\}},
\end{equation}
where $\mathbb{I}_{\{\cdot\}}$ is the indicator function.
This term primarily serves as a stabilizing signal, preventing malformed trajectories from dominating policy updates.

\emph{Contrastive Ranking Reward.}
A key challenge in training retrieval agents with reinforcement learning is the sparsity of
ranking supervision.
Binary correctness signals provide insufficient gradient information for list-wise
optimization, while standard positional penalties ignore the relative ordering between the
correct candidate and confusable hard negatives.
 
To address this, we propose a \textbf{Contrastive NDCG} (CNDCG) reward that provides
dense and discriminative feedback by explicitly modeling the relationship between the
correct candidate $j^*$ and hard negatives $\mathcal{H}$, defined as the top-$m$
non-correct candidates with the highest embedding similarity to the query
(formal definition in Appendix~\ref{app:cndcg}).
We assign graded relevance scores to each position $k$ in the predicted ranking $\hat{\pi}$:
\begin{equation}
    \mathrm{rel}_{\hat{\pi}(k)} =
    \begin{cases}
        2 & \hat{\pi}(k) = j^* \\
        1 & \hat{\pi}(k) \in \mathcal{H} \text{ and ranked after } j^* \\
        0 & \text{otherwise}
    \end{cases}
    \label{eq:cndcg_rel}
\end{equation}
Crucially, a hard negative receives $\mathrm{rel}=1$ \emph{only} when ranked below $j^*$;
ranking above $j^*$ yields zero credit, explicitly penalizing the most harmful errors.
The Discounted Cumulative Gain is computed over the top-$K_r$ positions as:
\begin{equation}
    \mathrm{DCG}(\hat{\pi}) = \sum_{k=1}^{K_r}
    \frac{2^{\,\mathrm{rel}_{\hat{\pi}(k)}}-1}{\log_2(k+1)},
    \label{eq:dcg}
\end{equation}
and the contrastive ranking reward is defined as:
\begin{equation}
    r_{\mathrm{rank}}(o_i) = \frac{\mathrm{DCG}(\hat{\pi})}{\mathrm{IDCG}},
    \label{eq:cndcg_reward}
\end{equation}
where $\mathrm{IDCG}$ is the ideal-ranking DCG with $j^*$ placed first
(see Appendix~\ref{app:cndcg}).
This formulation encourages the agent to
(i)~rank $j^*$ as high as possible,
(ii)~suppress hard negatives below $j^*$ for dense gradient feedback, and
(iii)~avoid placing any hard negative above $j^*$, which incurs zero reward.

\emph{Tool-Use Reward.}
The tool-use reward $r_{\text{tool}}$ directly governs the agent’s evidence acquisition behavior, encouraging visual inspection only when it contributes to correct decisions while discouraging redundant or excessive tool usage.
Let $N_{\text{tool}}$ denote the number of valid visual tool invocations in trajectory $o_i$, and let $k$ be the rank position of the correct candidate. We define:
\begin{equation}
\begin{aligned}
r_{\text{tool}}(o_i)
= \;&
\eta \cdot \mathbb{I}_{\{k = 1\}} \cdot \mathbb{I}_{\{N_{\text{tool}} > 0\}} \\
& - \rho \cdot \max(0, N_{\text{tool}} - \tau),
\end{aligned}
\end{equation}
where $\eta$ incentivizes successful evidence-based verification, $\rho$ penalizes excessive tool invocations, and $\tau$ specifies a tolerance threshold.
This design explicitly encodes the principle that \emph{effective} tool usage, rather than frequent usage, should be rewarded.

\emph{Policy Optimization.}
We instantiate EAPO using Group Relative Policy Optimization (GRPO) ~\cite{guo2025deepseek}. Given a group of $G$ trajectories sampled for the same query, we compute normalized advantages:
\begin{equation}
A_i = \frac{R_i - \mathrm{mean}(R)}{\mathrm{std}(R)}.
\end{equation}
Let $r_i(\theta) = \frac{\pi_\theta(o_i \mid q)}{\pi_{\theta_{\text{old}}}(o_i \mid q)}$, the final optimization objective is:

\begin{align}
\mathcal{J}_{\text{EAPO}}(\theta)
&= \mathbb{E}_{q}\left[
   \frac{1}{G} \sum_{i=1}^{G}
   \mathrm{clip}\!\left(r_i(\theta),\,1 \pm \epsilon\right) A_i
\right] \notag\\
&\quad - \lambda\, \mathbb{E}_{q}\!\left[
    \mathrm{KL}\!\left(\pi_\theta \,\Vert\, \pi_{\text{ref}}\right)
  \right]
\label{eq:eapo}
\end{align}

Through EAPO, the model learns not only \emph{what} to rank, but also \emph{how} and \emph{when} to acquire visual evidence in order to support reliable and efficient retrieval decisions.

\section{Experiments}

\subsection{Experimental Setup}
\label{experimental_setup}

\begin{table*}[t]
    \centering
    \caption{\footnotesize \textbf{Comparison with other methods on M-BEIR test set.} R@K refers to the Recall@K metric. $q^t$, $q^{i}$, $c^{t}$ and $c^{i}$ denote the text query, image query, text candidates and image candidates, respectively. Abbreviations: VN=VisualNews, F200K=Fashion200K, InfoS=InfoSeek, FIQ=FashionIQ. Best results in \textbf{bold}.}
    \renewcommand{\arraystretch}{1}
    \setlength{\tabcolsep}{2pt}
    \resizebox{\linewidth}{!}{
    \begin{tabular}{lccccccccccccccccc}%
    \toprule
     & \multicolumn{3}{c}{{$q^t \to c^i$}} & {$q^t \to c^t$} & \multicolumn{2}{c}{{$q^t \to (c^i, c^t)$}} & \multicolumn{3}{c}{{$q^i \to c^t$}} & {$q^i \to c^i$} & \multicolumn{2}{c}{{$(q^i, q^t) \to c^t$}} & \multicolumn{2}{c}{{$(q^i, q^t) \to c^i$}} & \multicolumn{2}{c}{{$(q^i, q^t) \to (c^i, c^t)$}} & \\
     \cmidrule(r){2-4} \cmidrule(r){5-5}  \cmidrule(r){6-7} \cmidrule(r){8-10} \cmidrule(r){11-11} \cmidrule(r){12-13} \cmidrule(r){14-15} \cmidrule(r){16-17}
     Models & VN  & COCO & F200K & WebQA & EDIS & WebQA & VN & COCO & F200K & NIGHTS & OVEN & InfoS & FIQ & CIRR & OVEN & InfoS & Avg. \\
    \cmidrule(r){2-4} \cmidrule(r){5-5}  \cmidrule(r){6-7} \cmidrule(r){8-10} \cmidrule(r){11-11} \cmidrule(r){12-13} \cmidrule(r){14-15} \cmidrule(r){16-17}
    & R@5 & R@5 & R@10 & R@5 & R@5 & R@5 & R@5 & R@5 & R@10 & R@5 & R@5 & R@5 & R@10 & R@5 & R@5 & R@5 & \\
    \midrule
    CLIP-L~\cite{radford2021learning} & 43.3 & 61.1 & 6.6 & 36.2 & 43.3 & 45.1 & 41.3 & 79.0  & 7.7 & 26.1 & 24.2 & 20.5 & 7.0 & 13.2 & 38.8 & 26.4 & 32.5    \\
    SigLIP~\cite{zhai2023sigmoid} & 30.1 & 75.7 & 36.5 & 39.8 & 27.0 & 43.5 & 30.8 & 88.2  & 34.2 & 28.9 & 29.7 & 25.1 & 14.4 & 22.7 & 41.7 & 27.4 & 37.2  \\
    BLIP~\cite{li2022blip} & 16.4 & 74.4 & 15.9 & 44.9 & 26.8 & 20.3 & 17.2 & 83.2  & 19.9 & 27.4 & 16.1 & 10.2 & 2.3 & 10.6 & 27.4 & 16.6 & 26.8  \\
    BLIP2~\cite{li2023blip} & 16.7 & 63.8 & 14.0 & 38.6 & 26.9 & 24.5 & 15.0 & 80.0  & 14.2 & 25.4 & 12.2 & 5.5 & 4.4 & 11.8 & 27.3 & 15.8 & 24.8  \\
    $\text{UniIR-BLIP}_{\text{FF}}$~\cite{wei2024uniir} & 23.4 & 79.7 & 26.1 & 80.0 & 50.9 & 79.8 & 22.8 & 89.9 & 28.9 & 33.0 & 41.0 & 22.4 & 29.2 & 52.2 & 55.8 & 33.0 & 46.8  \\
    $\text{UniIR-CLIP}_{\text{SF}}$~\cite{wei2024uniir} & 42.6 & 81.1 & 18.0 & 84.7 & 59.4 & 78.7 & 43.1 & 92.3 & 18.3 & 32.0 & 45.5 & 27.9 & 24.4 & 44.6 & 67.6 & 48.9 & 50.6  \\
    Qwen2.5-VL-3B~\cite{bai2025qwen2} & 36.0 & 67.8 & 16.1 & 69.5 & 45.2 & 61.7 & 23.3 & 82.3 & 12.0 & 20.9 & 36.7 & 22.3 & 24.3 & 53.5 & 56.4 & 49.8 & 42.4 \\
    Qwen2.5-VL-7B~\cite{bai2025qwen2} & 40.2 & 71.9 & 20.3 & 71.9 & 49.4 & 64.5 & 29.3 & 84.6 & 19.4 & 25.5 & 42.4 & 32.1 & 25.0 & 55.1 & 60.8 & 54.9 & 46.7 \\
    Vision-R1-7B~\cite{huang2025vision}& 41.9 & 75.0 & 22.0 & 70.6 & 51.3 & 69.1 & 35.4 & 85.1 & 22.4 & 25.9 & 48.8 & 44.0 & 29.2 & 57.7 & 66.2 & 59.0 & 50.2 \\
    VLM-R1-7B~\cite{shen2025vlm}& 40.5 & 77.2 & 22.5 & 72.3 & 50.0 & 67.9 & 36.2 & 86.3 & 20.9 & 26.4 & 48.8 & 37.5 & 29.9 & 57.4 & 64.0 & 62.3 & 50.0 \\
    DeepEyes-7B~\cite{zheng2025deepeyes} & 43.5 & 73.2 & 23.5 & 75.4 & 55.2 & 70.1 & 38.6 & 86.5 & 24.1 & 28.3 & 50.4 & 46.2 & 31.5 & 59.2 & 68.4 & 63.2 & 52.3 \\
    MM-Embed-7B~\cite{lin2024mm}& 41.0 & 71.3 & 17.1 & 95.9 & 68.8 & 85.0 & 41.3 & 90.1 & 18.4 & 32.4 & 42.1 & 42.3 & 25.7 & 50.0 & 64.1 & 57.7 & 52.7 \\
    LamRA-7B~\cite{liu2024lamra} & 48.0 & 85.2 & 32.9 & 96.7 & 75.8 & 87.7 & 48.6 & 92.3 & 36.1 & 33.5 & 59.2 & 64.1 & 37.8 & 63.3 & 79.2 & 78.3 & 63.7 \\
    U-MARVEL-7B~\cite{li2025u} & 49.4 & 85.6 & 34.2 & \textbf{98.5} & 81.4 & 89.4 & 50.5 & 88.4 & 37.7 & 34.7 & 63.7 & 62.9 & 38.2 & 63.2 & 80.8 & 78.9 & 64.8 \\
    Retrv-R1-3B~\cite{zhu2025retrv} & 46.6 & 83.0 & 34.4 & 96.7 & 77.9 & 88.2 & 48.0 & 92.7 & 34.5 & 34.7 & 64.3 & 70.0 & 45.2 & 67.9 & 83.3 & 80.0 & 65.5 \\
    Retrv-R1-7B~\cite{zhu2025retrv} & 50.5 & 86.7 & 39.3 & 97.4 & 82.1 & 89.5 & 51.4 & 94.1 & \textbf{39.8} & 39.5 & 69.0 & \textbf{75.6} & 49.5 & 72.3 & 86.6 & 83.7 & 69.2 \\
    \midrule
    \rowcolor{gray!20}
     V-Retrver-7B & \textbf{51.8} & \textbf{87.5} & \textbf{40.3} & 96.9 & \textbf{82.9} & \textbf{90.2} & \textbf{52.2} & \textbf{94.8} & 37.8 & \textbf{39.8} & \textbf{69.8} & 73.2 & \textbf{51.2} & \textbf{73.5} & \textbf{87.8} & \textbf{85.0} & \textbf{69.7} \\
    \bottomrule
     \end{tabular}
     }
    \vspace{-1pt}
    \label{table_main_results}
    \end{table*}

\begin{table*}[t!]
\centering
\begin{minipage}[t]{0.49\linewidth}
    \centering
    \caption{\footnotesize \textbf{Experimental results on unseen datasets.} $q^{\text{dialog}}$ and $(q^i \oplus q^t)$ refer to dialog queries and multi-interleaved image-text queries, respectively.}
    \label{table_unseen_datasets}
    \setlength{\tabcolsep}{2pt}
    \resizebox{\linewidth}{!}{
        \begin{tabular}{lccccc}
        \toprule
         & \multicolumn{2}{c}{{$(q^i, q^t) \to c^i$}} & \multicolumn{1}{c}{{$q^{\text{dialog}} \to c^i$}} & \multicolumn{2}{c}{{$(q^i \oplus q^t) \to c^i$}} \\
        \cmidrule(r){2-3} \cmidrule(r){4-4}  \cmidrule(r){5-6}
        Models & CIRCO & GeneCIS & VisD & VIST & MT-FIQ \\
        \cmidrule(r){2-3} \cmidrule(r){4-4}  \cmidrule(r){5-6}
         & MAP@5 & R@1 & R@1 & R@1 & R@5 \\
        \midrule
        CLIP-L~\cite{radford2021learning} & 4.0 & 13.3 & 23.7 & 0.6 & 17.7\\
        UniIR-CLIP~\cite{wei2024uniir} & 12.5 & 16.8 & 26.8 & 0.6 & 39.4 \\
        E5-V~\cite{jiang2024e5} & 24.8 & 18.5 & 54.6 & 10.0 & 19.2 \\
        MagicLens-L~\cite{zhang2024magiclens} & 29.6 & 16.3 & 28.0 & 3.3 & 22.6  \\
        MM-Embed-7B \cite{lin2024mm} & 35.5 & 22.9 & 64.7 & 25.7 & 59.0 \\
        LamRA-7B \cite{liu2024lamra} & 42.8 & 24.8 & 70.9 & 28.6 & 63.9\\
        \midrule
        \rowcolor{gray!20} V-Retrver-7B & \textbf{48.2} & \textbf{30.7} & \textbf{75.1} & \textbf{31.2} & \textbf{68.3} \\
        \bottomrule
        \end{tabular}
    }
\end{minipage}
\hfill
\begin{minipage}[t]{0.49\linewidth}
    \centering
    \caption{\footnotesize \textbf{Experimental results on held-out tasks.} $^*$ indicates training on remaining tasks only, w/o exposure to the three held-out tasks.}
    \label{table_held_out_datasets}
    \setlength{\tabcolsep}{2pt}
    \resizebox{\linewidth}{!}{
        \begin{tabular}{lcccccc}
        \toprule
         & {$q^i \to c^i$} & \multicolumn{2}{c}{{$(q^i, q^t) \to c^t$}} & \multicolumn{2}{c}{{$(q^i, q^t) \to (c^i, c^t)$}}\\
         \cmidrule(r){2-2} \cmidrule(r){3-4}  \cmidrule(r){5-6}
         Models & NIGHTS & OVEN & InfoS & OVEN & InfoS & Avg. \\
         \cmidrule(r){2-2} \cmidrule(r){3-4}  \cmidrule(r){5-6}
        & R@5 & R@5 & R@5 & R@5 & R@5 & \\
        \midrule
        \multicolumn{7}{l}{\color{gray}{\textit{Supervised}}} \\
        \addlinespace[0.15em]
        $\text{UniIR-BLIP}_{\text{FF}}$~\cite{wei2024uniir} & 33.0 & 41.0 & 22.4 & 55.8 & 33.0 & 37.0 \\
        $\text{UniIR-CLIP}_{\text{SF}}$~\cite{wei2024uniir} & 32.0 & 45.5 & 27.9 & 67.6 & 48.9 & 44.4 \\
        \midrule
        \multicolumn{7}{l}{\color{gray}{\textit{Zero-shot}}} \\
        \addlinespace[0.15em]
        Qwen2.5-VL-7B \cite{bai2025qwen2} & 20.3 & 38.5 & 40.4 & 53.6 & 44.9 & 39.5\\
        Vision-R1-7B \cite{huang2025vision}& 22.9 & 39.8 & 42.9 & 57.4 & 46.5 & 41.9\\
        LamRA-7B$^*$ \cite{liu2024lamra} & 29.2 & 46.9 & 54.2 & 65.1 & 59.1 & 50.9 \\
        \rowcolor{gray!20} V-Retrver-7B$^*$ & \textbf{36.2} & \textbf{57.8} & \textbf{65.9} & \textbf{75.3} & \textbf{70.3} & \textbf{61.1}\\
        \bottomrule
        \end{tabular}
    }
    \vspace{-2em}
\end{minipage}
\end{table*}

\paragraph{Datasets and Metrics.}
We train V-Retrver on M-BEIR~\citep{wei2024uniir}, which encompasses eight distinct
retrieval tasks across 10 datasets comprising 1.1M training samples.
We evaluate on the M-BEIR test set and five previously unseen datasets:
CIRCO~\citep{baldrati2023zero}, GeneCIS~\citep{vaze2023genecis},
Visual Storytelling~\citep{huang2016visual}, Visual Dialog~\citep{das2017visual},
and Multi-round FashionIQ~\citep{yuan2021conversational}.
We primarily use Recall@$K$ as the evaluation metric and additionally report
MAP@5 for CIRCO. Dataset statistics and evaluation settings are in Appendix~\ref{app:datasets}.

\paragraph{Baselines.}
We compare V-Retrver against four categories:
(1)~foundational VLMs (CLIP, SigLIP, BLIP, BLIP2);
(2)~fine-tuned universal retrievers (UniIR~\citep{wei2024uniir}, MM-Embed~\citep{lin2024mm}, LamRA~\citep{liu2024lamra}, U-MARVEL~\citep{li2025u});
(3)~reasoning-enhanced models (Vision-R1~\citep{huang2025vision}, VLM-R1~\citep{shen2025vlm}, Retrv-R1~\citep{zhu2025retrv}); and
(4)~agentic tool-use models (DeepEyes~\citep{zheng2025deepeyes}).
Detailed baseline descriptions and evaluation protocols are in Appendix~\ref{app:baselines}.

\paragraph{Reranking Setup.}
Following the coarse-to-fine paradigm, the embedding model $\phi$ first retrieves the top-$K$ candidates, over which the reasoning agent $\theta$ performs listwise reranking in a single inference pass.
We set $K{=}50$ for M-BEIR evaluation (reranking from local pool) and $K{=}10$ for unseen datasets.

\paragraph{Implementation Details.}
Our model is initialized from Qwen2.5-VL-7B-Instruct~\citep{bai2025qwen2}.
The vision encoder remains frozen throughout all training stages while the language model is fine-tuned.
Complete training configurations and hyperparameter settings are in Appendix~\ref{app:impl}.

\subsection{Main Results}

\noindent\textbf{Performance on M-BEIR.}
As presented in Table~\ref{table_main_results}, V-Retrver-7B establishes a new state-of-the-art on M-BEIR with an average Recall of 69.7\%, surpassing U-MARVEL-7B by +4.9\% (64.8\%) and Retrv-R1-7B by +0.5\% (69.2\%). Gains are especially pronounced on fine-grained visual tasks: V-Retrver achieves 51.2\% on FIQ and 73.5\% on CIRR, outperforming U-MARVEL-7B (38.2\% and 63.2\%) by large margins. While Retrv-R1-7B surpasses V-Retrver on certain text-centric subtasks (e.g., F200K $q^i{\to}c^t$, InfoS $(q^i,q^t){\to}c^t$), V-Retrver achieves consistent advantages on visually ambiguous tasks that require fine-grained inspection.

\vspace{2pt}\noindent\textbf{Generalization to Unseen Datasets.}
Table~\ref{table_unseen_datasets} confirms robustness on datasets not seen during training. V-Retrver consistently outperforms specialized models and generalist MLLMs, achieving MAP@5 of 48.2 on CIRCO (vs.\ MM-Embed 35.5 and LamRA 42.8) and R@1 of 30.7 on GeneCIS (vs.\ 24.8 for LamRA).

\vspace{2pt}\noindent\textbf{Robustness on Held-out Tasks.}
Even without exposure to three modality-combination types during training, V-Retrver$^*$ achieves 61.1\% average Recall (Table~\ref{table_held_out_datasets}), outperforming LamRA-7B$^*$ (50.9\%) by +10.2\%. This confirms that MIER effectively decouples evidence-gathering reasoning from specific input modalities.

\subsection{Ablation Study}

\begin{figure*}[t]
    \centering
    \begin{subfigure}[b]{0.32\textwidth}
        \centering
        \includegraphics[width=\textwidth]{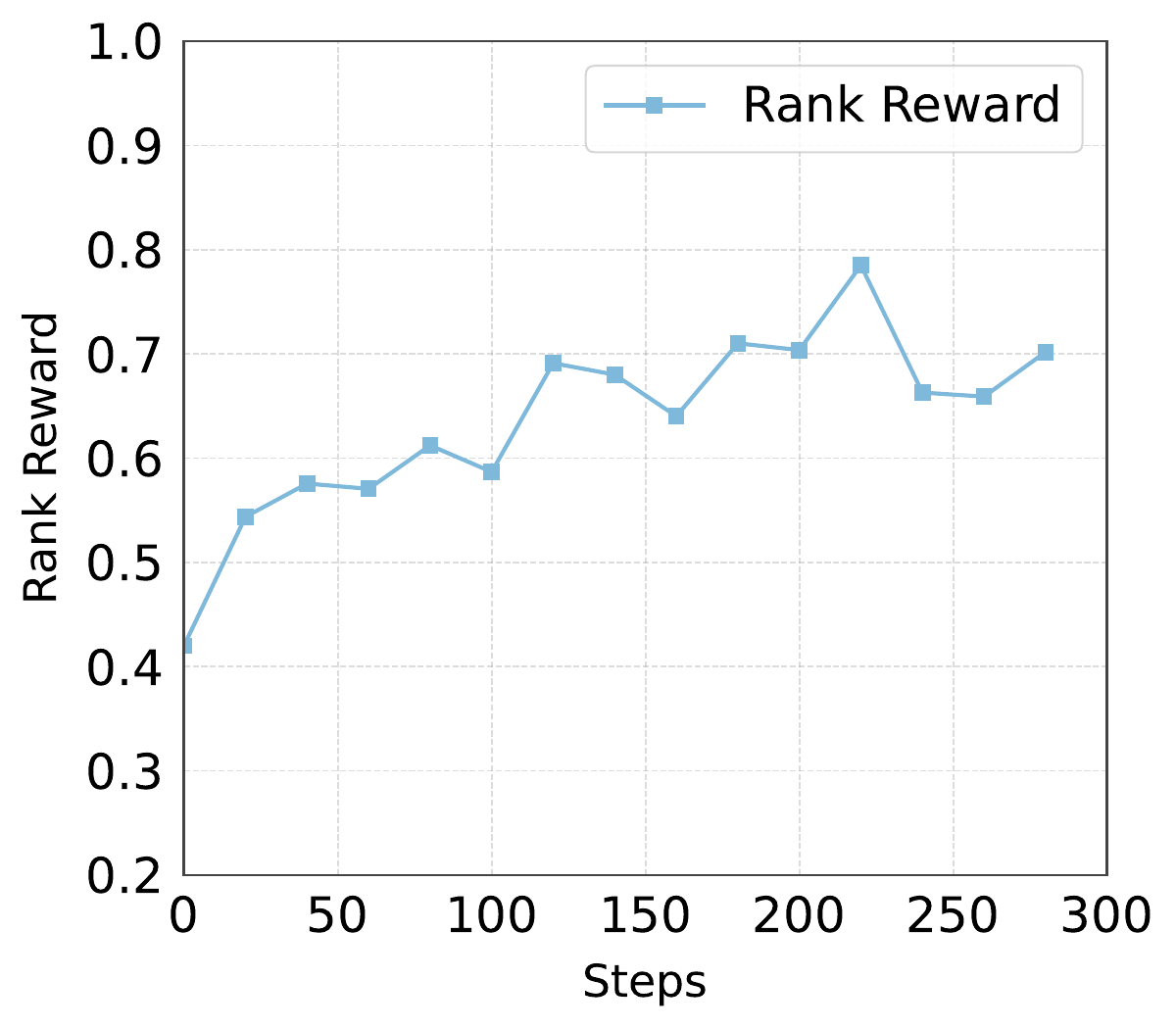}
        \vspace{-1.8em}
        \caption{Rank Reward}
        \label{fig:curve_rank}
    \end{subfigure}
    \hfill
    \begin{subfigure}[b]{0.32\textwidth}
        \centering
        \includegraphics[width=\textwidth]{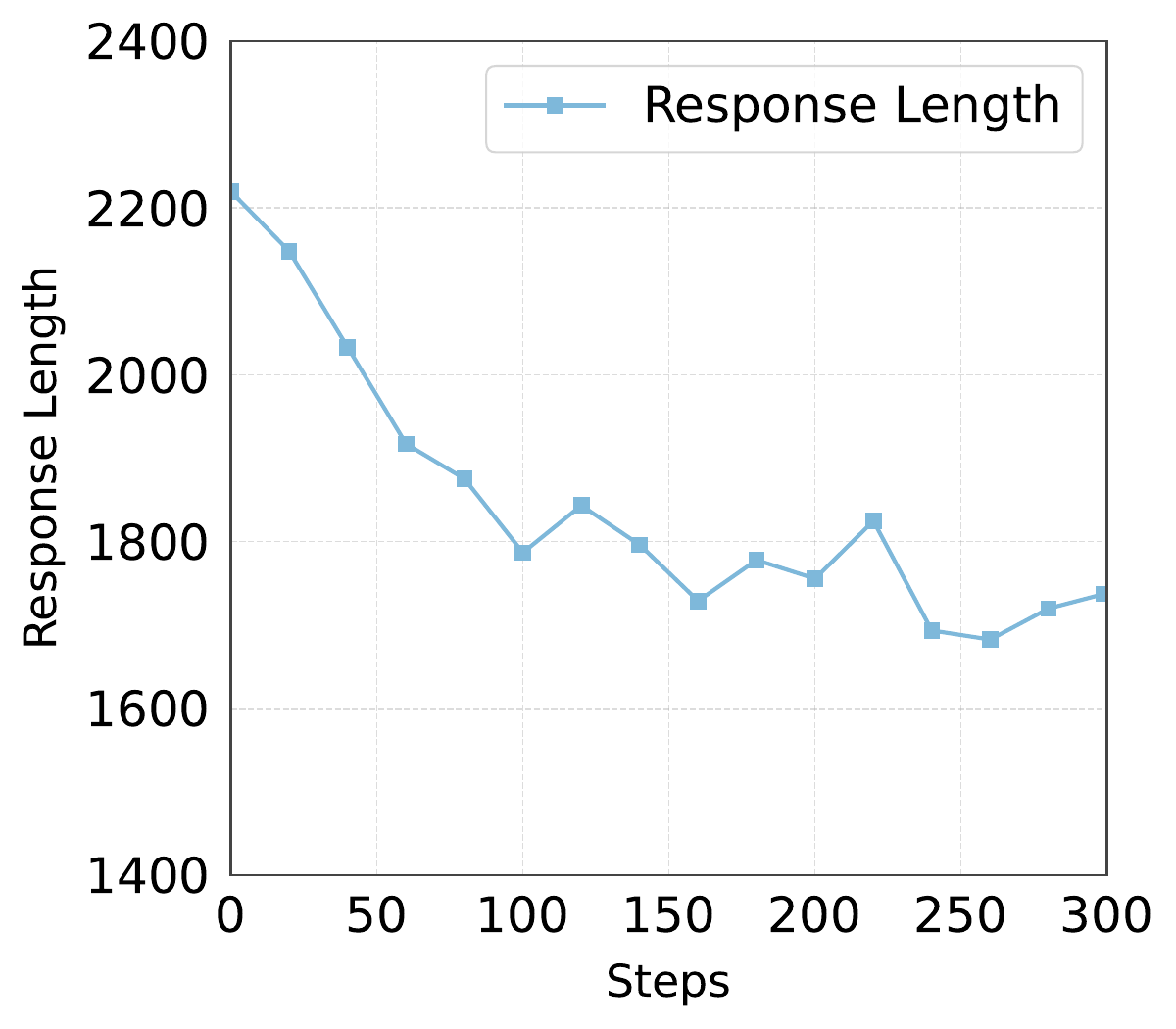}
        \vspace{-1.8em}
        \caption{Response Length}
        \label{fig:curve_length}
    \end{subfigure}
    \hfill
    \begin{subfigure}[b]{0.32\textwidth}
        \centering
        \includegraphics[width=\textwidth]{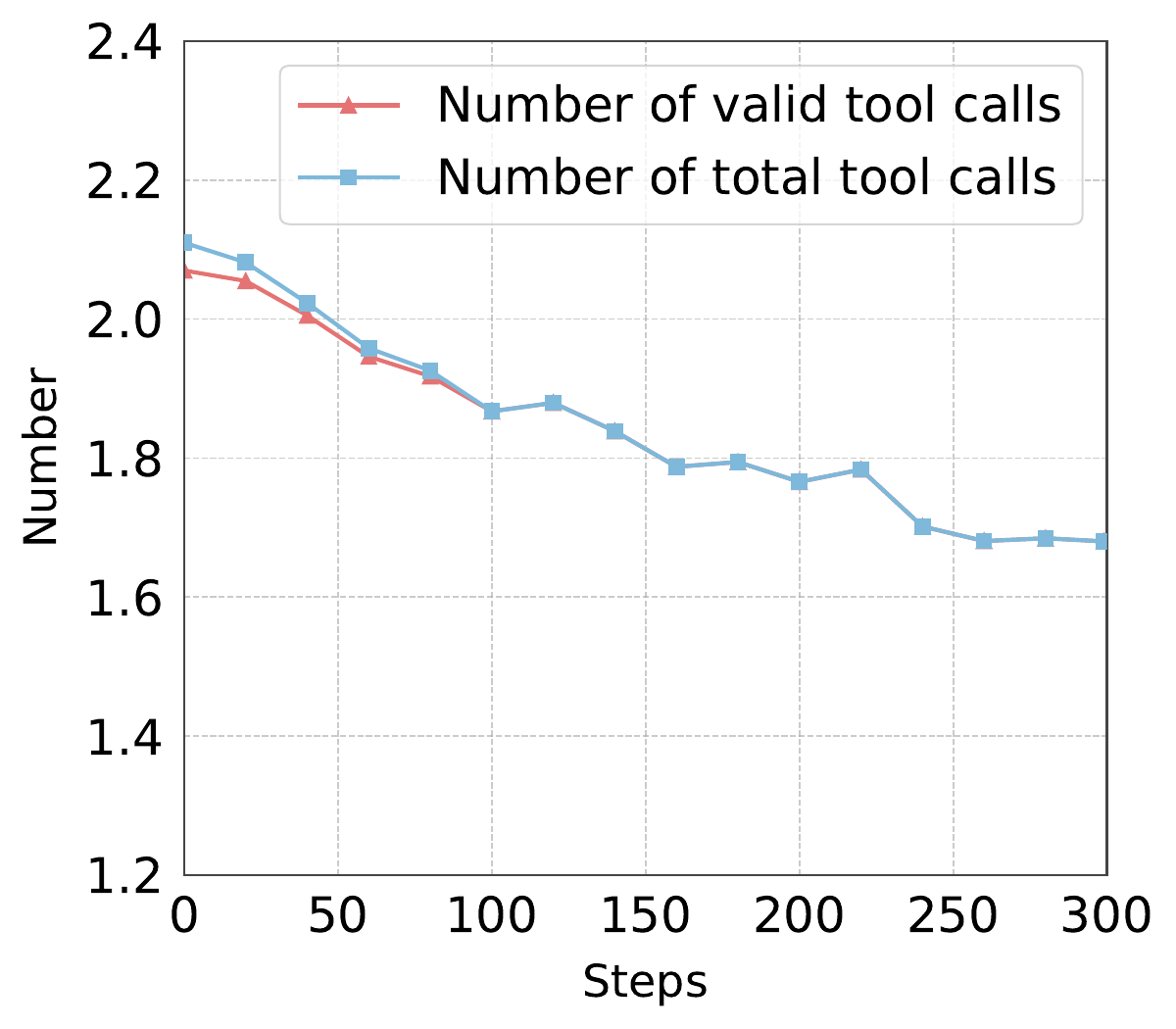}
        \vspace{-1.8em}
        \caption{Tool Calls}
        \label{fig:curve_tools}
    \end{subfigure}
    \caption{RL training curves showing rank reward improvement, response length stabilization, and convergence of valid vs.\ total tool calls.}
    \label{fig:training_curves}
    \vspace{0.6em}
    \begin{minipage}[t]{0.48\textwidth}
        \centering
        \captionsetup{type=table}
        \caption{\footnotesize \textbf{Ablation: visual tool-use mechanism.} Multimodal interleaved CoT (with Visual Tool) vs.\ text-only baseline under identical RL training.}
        \label{table_ablation_tool}
        \vskip 0.04in
        \setlength{\tabcolsep}{3pt}
        \resizebox{\linewidth}{!}{
        \begin{tabular}{lccccc}
        \toprule
         & $q^t \to c^i$ & $q^i \to c^t$ & $(q^i, q^t) \to c^i$ & $(q^i, q^t) \to c^t$ &  \\
        \cmidrule(lr){2-2} \cmidrule(lr){3-3} \cmidrule(lr){4-4} \cmidrule(lr){5-5}
         Variants & COCO & F200K & CIRR & OVEN & Avg.\\
         \cmidrule(lr){2-2} \cmidrule(lr){3-3} \cmidrule(lr){4-4} \cmidrule(lr){5-5}
         & R@5 & R@10 & R@5 & R@5 & \\
        \midrule
         Qwen2.5-VL-7B \cite{bai2025qwen2}& 71.9 & 19.4 & 55.1 & 42.4 & 47.2 \\
         RL w/o tool & 84.1 & 33.2 & 66.5 & 63.2 & 61.8 \\
        \midrule
        \rowcolor{gray!20} V-Retrver-7B & \textbf{87.5} & \textbf{37.8} & \textbf{73.5} & \textbf{69.8} & \textbf{67.2} \\
        \bottomrule
        \end{tabular}
        }
    \end{minipage}
    \hfill
    \begin{minipage}[t]{0.48\textwidth}
        \centering
        \captionsetup{type=table}
        \caption{\footnotesize \textbf{Ablation: training stages.} Impact of Cold Start (SFT), Rejection Sampling Fine-Tuning (RSFT), and Reinforcement Learning (RL).}
        \label{table_ablation_study}
        \vskip 0.04in
        \setlength{\tabcolsep}{3pt}
        \resizebox{\linewidth}{!}{
        \begin{tabular}{lccccc}
        \toprule
         & $q^t \to c^i$ & $q^i \to c^t$ & $(q^i, q^t) \to c^i$ & $(q^i, q^t) \to c^t$ &  \\
        \cmidrule(lr){2-2} \cmidrule(lr){3-3} \cmidrule(lr){4-4} \cmidrule(lr){5-5}
         Training Stage& COCO & F200K & CIRR & OVEN & Avg.\\
         \cmidrule(lr){2-2} \cmidrule(lr){3-3} \cmidrule(lr){4-4} \cmidrule(lr){5-5}
         & R@5 & R@10 & R@5 & R@5 & \\
        \midrule
         Qwen2.5-VL-7B \cite{bai2025qwen2}& 71.9 & 19.4 & 55.1 & 42.4 & 47.2 \\
         w/o SFT \& RSFT \& RL & 71.5 & 18.1 & 53.4 & 40.2 & 45.8 \\
         w/o RSFT \& RL & 83.2 & 31.6 & 63.7 & 59.0 & 59.4 \\
         w/o RSFT & 87.2 & 37.3 & 72.4 & 68.3 & 66.3 \\
         w/o RL & 83.9 & 32.8 & 65.3 & 61.5 & 60.9 \\
        \midrule
        \rowcolor{gray!20} V-Retrver-7B & \textbf{87.5} & \textbf{37.8} & \textbf{73.5} & \textbf{69.8} & \textbf{67.2} \\
        \bottomrule
        \end{tabular}
        }
    \end{minipage}
    \vspace{-8pt}
\end{figure*}

\noindent\textbf{Impact of Training Stages.}
Table~\ref{table_ablation_study} ablates each training stage. Zero-shot tool invocation without any alignment collapses to 45.8\% (below the 47.2\% baseline), confirming alignment is critical. SFT alone reaches 59.4\%; adding RL without RSFT gives 66.3\%; SFT+RSFT without RL yields 60.9\%. The full three-stage pipeline achieves 67.2\%, demonstrating that each stage addresses specific shortcomings of the previous one.

\noindent\textbf{Effectiveness of Visual Tool.}
Table~\ref{table_ablation_tool} compares V-Retrver against a text-only RL variant trained identically but without visual tools. The text-only variant achieves 61.8\% vs.\ V-Retrver's 67.2\%, confirming that active zoom-in and image selection resolve fine-grained ambiguities that compressed static representations cannot capture. The baseline frequently struggles to distinguish between hard negatives that share identical global semantics but differ only in minute visual details.

\subsection{Analysis}

\noindent\textbf{RL Training Dynamics.}
Fig.~\ref{fig:training_curves} shows that rank reward improves steadily under EAPO. Initially, valid tool calls are slightly fewer than total invocations, reflecting occasional formatting errors from early SFT/RSFT training. As RL progresses, the two curves converge, confirming that policy optimization eliminates erroneous tool actions. Concurrently, response length and tool-call frequency decrease and stabilize, demonstrating that the model learns to invoke tools only when necessary, with an average of 1.6 to 1.7 calls per query.

\noindent\textbf{Efficiency Analysis.}
V-Retrver employs a listwise ranking approach, preserving uncompressed visual representations while remaining competitive in latency. Against the pointwise baseline LamRA-Rank that invokes the LLM once per candidate (50 calls for $K{=}50$, ITR $4.41\times$), V-Retrver is substantially faster. Compared to Retrv-R1, which achieves lower ITR by compressing visual tokens, V-Retrver accepts a modest latency overhead in exchange for fine-grained visual fidelity and higher retrieval accuracy. Full inference time ratio (ITR) and GPU memory ratio (GMR) comparisons are in Appendix~\ref{app:efficiency}.

\section{Conclusion}

In this paper, we presented V-Retrver,  an evidence-driven MLLM
framework tailored for universal multimodal retrieval. 
V-Retrver adopts multimodal interleaved Chain-of-Thought (CoT) reasoning, enabling the model to dynamically inspect and verify candidate images through visual tool invocation, thereby achieving more fine-grained ranking of candidate result lists.
We adopt a three-stage training pipeline to activate and reinforce multimodal interleaved CoT reasoning abilities.
Extensive experimental results demonstrate that V-Retrver achieves significant improvements in both model effectiveness and task generalization.
We regard V-Retrver to be an important step toward effectively introducing agentic MLLMs to enhance downstream multimodal tasks, laying a solid foundation for building general agentic MLLMs with advanced reasoning capabilities. In future work, we plan to extend V-Retrver by exploring fine-grained token-level advantage estimation. Integrating nuanced generalized advantage estimation across the entire multi-turn reasoning sequence could effectively resolve credit assignment challenges, ultimately driving further optimizations in real-time inference efficiency and broader tool-use scalability.

% \section*{Impact Statement}
% This paper presents work whose goal is to advance the field of machine learning. There are many potential societal consequences of our work, none of which we feel must be specifically highlighted here.

\section*{Limitations}

Despite its strong performance, V-Retrver still has several limitations. First, the multi-turn tool invocation process introduces a credit assignment challenge: since EAPO evaluates trajectories at the sequence level, it is difficult to attribute the contribution of each individual tool call to the final ranking outcome, particularly in long reasoning chains. Fine-grained step-level credit assignment remains an open problem for agentic retrieval training. Second, the inference efficiency of V-Retrver remains a concern for large-scale deployment. The multi-turn agentic reasoning with interleaved tool invocations and uncompressed visual representations incurs non-negligible latency and memory overhead, limiting applicability in latency-sensitive scenarios with large candidate pools. Future work will explore step-level credit assignment mechanisms, lightweight adaptive inference strategies such as early-exit and dynamic tool scheduling, and extensions to broader downstream tasks such as multimodal recommendation and retrieval-augmented generation.

\section*{Ethics Statement}

This work focuses on universal multimodal retrieval and does not involve the collection or annotation of human subjects data. All datasets used in our experiments are publicly available and were obtained in accordance with their respective licenses. The proposed framework, V-Retrver, is intended to improve the accuracy and reliability of multimodal retrieval systems. We do not foresee direct negative societal consequences arising from this research, though we acknowledge that any retrieval system may inherit biases present in its training data.

\section*{Acknowledgments}

This work is supported by the Natural Science Foundation of Guangdong Province (2025A1515011946) and the Basic and Frontier Research Project of PCL (2025QYA001).

\bibliography{custom}

\appendix

% ================================================================
% APPENDIX — revised layout
% Requirements:
%   • Experimental sections first (Efficiency, Ablations, RAG, Impl.)
%   • Table 6,10,11,13 → table*  (double-column)
%   • Table 7,8,9,12  → table   (single-column)
%   • Uniform \small font for all tables
%   • No overflow; \resizebox used where needed
%   • All em/en dashes in prose replaced with words
%   • Cross-references added where missing
%   • Float placement [htbp] to keep tables near their text
%   • Qualitative figures kept as [p] for clean page layout
% ================================================================

\clearpage

% ----------------------------------------------------------------
% SECTION 1 — Efficiency Analysis  (Table 6, double-column)
% ----------------------------------------------------------------
\section{Efficiency Analysis}
\label{app:efficiency}

We evaluate V-Retrver's inference efficiency against strong baselines
using two metrics: \textbf{ITR} (Inference Time Ratio, relative to
V-Retrver) and \textbf{GMR} (GPU Memory Ratio, relative to V-Retrver).
Results under two candidate pool sizes, $K{=}20$ and $K{=}50$, are
reported in Table~\ref{tab:efficiency}, together with the representative
CIRR R@5 metric to cover compositional retrieval settings.

% ----------------------------------------------------------------
% Table 6 — double-column，去掉 resizebox，用 tabcolsep 控宽
% ----------------------------------------------------------------
\begin{table*}[htbp]
\centering
\small
\setlength{\tabcolsep}{5pt}
\caption{Efficiency comparison. ITR and GMR are relative to
  V-Retrver ($=1$); $\downarrow$ is better. All models use a
  \emph{single} MLLM inference call, except LamRA-Rank-Pointwise
  which invokes the LLM independently for each candidate.}
\label{tab:efficiency}
\begin{tabular}{lccc|ccc}
\toprule
 & \multicolumn{6}{c}{$(q_i, q_t) \rightarrow c_i$ (CIRR)} \\
\cmidrule{2-7}
 & \multicolumn{3}{c|}{\textbf{$K = 20$}}
 & \multicolumn{3}{c}{\textbf{$K = 50$}} \\
\cmidrule(r){2-4}\cmidrule(l){5-7}
Methods & R@5 & ITR~$\downarrow$ & GMR~$\downarrow$
        & R@5 & ITR~$\downarrow$ & GMR~$\downarrow$ \\
\midrule
Qwen2.5-VL-7B           & 52.6 & 0.60 & 0.71 & 55.1 & 0.62 & 0.70 \\
Vision-R1-7B            & 53.7 & 0.94 & 0.92 & 57.7 & 0.95 & 0.93 \\
VLM-R1-7B               & 53.1 & 0.96 & 0.95 & 57.4 & 0.96 & 0.96 \\
Retrv-R1-7B             & 69.9 & 0.25 & 0.31 & 72.3 & 0.22 & 0.28 \\
LamRA-Rank-Pointwise-7B & 65.4 & 4.17 & 0.21 & 67.9 & 4.41 & 0.11 \\
\midrule
\rowcolor{gray!20}
V-Retrver-7B & \textbf{71.2} & 1.00 & 1.00
             & \textbf{73.5} & 1.00 & 1.00 \\
\bottomrule
\end{tabular}
\end{table*}

\paragraph{Comparison with pointwise methods.}
LamRA-Rank-Pointwise invokes the LLM \emph{independently for each
candidate}: processing $K$ candidates requires $K$ separate MLLM
calls, each containing only a single candidate.  This pointwise
paradigm leads to an ITR of $4.41\times$ at $K{=}50$, as inference
latency scales linearly with the candidate pool size.  Moreover,
because each call processes only one candidate in isolation, the model
lacks a global receptive field and cannot directly contrast similar
candidates, which is a critical weakness for distinguishing visually
similar hard negatives.

\paragraph{Comparison with token-compression methods.}
Retrv-R1 achieves lower ITR by heavily compressing visual tokens,
allowing all 50 candidates to fit within a single inference pass.
However, this compression inevitably discards fine-grained visual
details.  V-Retrver accepts a modest latency overhead (ITR $1.0\times$
vs.\ Retrv-R1's $0.22\times$ at $K{=}50$) to preserve uncompressed
visual representations, enabling precise tool-based inspection and
yielding $+1.2\%$ on CIRR at $K{=}50$.

\paragraph{Tool-call efficiency under EAPO.}
As shown in Fig.~\ref{fig:training_curves}(c), the model stabilizes at
\textbf{1.6 to 1.7 average tool calls per query}, having learned under
EAPO to invoke visual inspection only when it genuinely resolves
ambiguity.  This autonomously acquired efficiency makes V-Retrver
well-suited for precision-critical applications (e.g., medical imaging,
trademark detection) where retrieval accuracy outweighs minor latency
overhead.

% ----------------------------------------------------------------
% SECTION 2 — Reward Component Ablation  (Table 7, single-column)
% ----------------------------------------------------------------
\section{Reward Component Ablation}
\label{app:reward_ablation}

To validate each component of the EAPO reward (Eq.~1), we train three
variants by removing one term at a time, keeping all other settings
identical.  Results are presented in Table~\ref{tab:reward_ablation}.

% ----------------------------------------------------------------
% Table 7 — 改为 double-column，字体与 Table 6 一致
% ----------------------------------------------------------------
\begin{table*}[htbp]
\centering
\small
\setlength{\tabcolsep}{12pt}
\caption{Ablation of EAPO reward components. Removing any term
  degrades performance; $r_{\text{rank}}$ has the largest individual
  impact.}
\label{tab:reward_ablation}
\begin{tabular}{lccccc}
\toprule
Variants & COCO R@5 & F200K R@10 & CIRR R@5 & OVEN R@5 & Avg. \\
\midrule
\rowcolor{gray!20}
V-Retrver-7B (full)      & \textbf{87.5} & \textbf{37.8}
                          & \textbf{73.5} & \textbf{69.8} & \textbf{67.2} \\
w/o $r_{\text{format}}$  & 86.2 & 36.1 & 71.5 & 68.2 & 65.5 \\
w/o $r_{\text{rank}}$    & 85.5 & 35.3 & 70.4 & 68.0 & 64.8 \\
w/o $r_{\text{tool}}$    & 86.8 & 37.0 & 72.1 & 68.5 & 66.1 \\
\bottomrule
\end{tabular}
\end{table*}

Removing $r_{\text{format}}$ leads to a 1.7\% drop in average recall,
as malformed trajectories pollute policy updates.  Replacing
$r_{\text{rank}}$ with a binary correctness signal causes the largest
performance drop (2.4\%), confirming the critical value of dense
gradient feedback from CNDCG.  Finally, removing $r_{\text{tool}}$
reduces performance by 1.1\%, as the model loses the incentive to
perform successful evidence-based verification and has no mechanism to
suppress redundant calls.

% ----------------------------------------------------------------
% SECTION 3 — Tool Tolerance Threshold Ablation (Table 8, single-col)
% ----------------------------------------------------------------
\section{Tool Tolerance Threshold Ablation}
\label{app:tau_ablation}

We ablate the tolerance threshold $\tau$ in the tool-use reward
(Eq.~6) against a dynamic variant that scales $\tau$ by the semantic
similarity variance of the top-$K$ candidates.  Results are
summarized in Table~\ref{tab:tau_ablation}.
% ----------------------------------------------------------------
% Table 8 — 改为 double-column，字体与 Table 6 一致
% ----------------------------------------------------------------
\begin{table*}[htbp]
\centering
\small
\setlength{\tabcolsep}{10pt}
\caption{Ablation of tool tolerance threshold $\tau$. Fixed
  $\tau{=}1$ achieves the best accuracy and efficiency trade-off.}
\label{tab:tau_ablation}
\begin{tabular}{lcccccc}
\toprule
Variants & COCO R@5 & F200K R@10 & CIRR R@5 & OVEN R@5 & Avg. & Avg.\ Calls \\
\midrule
Fixed $\tau{=}0$
  & 87.1 & 37.2 & 73.2 & 69.5 & 66.8 & 1.12 \\
\rowcolor{gray!20}
Fixed $\tau{=}1$ (ours)
  & \textbf{87.5} & 37.8 & 73.5 & 69.8 & \textbf{67.2} & 1.68 \\
Fixed $\tau{=}2$
  & 87.8 & \textbf{38.2} & \textbf{73.9} & \textbf{70.2} & 67.5 & 2.45 \\
Dynamic $\tau$
  & 87.8 & 37.8 & 73.6 & 70.0 & 67.3 & 2.15 \\
\bottomrule
\end{tabular}
\end{table*}

Stricter $\tau{=}0$ reduces accuracy by discouraging necessary tool
calls.  More lenient $\tau{=}2$ or dynamic $\tau$ yield marginal
accuracy gains ($\leq 0.3\%$) while inflating overhead (+28\% to
+46\% more calls relative to $\tau{=}1$).  Fixed $\tau{=}1$ achieves
the optimal accuracy and efficiency balance for scalable retrieval.

% ----------------------------------------------------------------
% SECTION 4 — RAG Applications  (Table 13, double-column)
% ----------------------------------------------------------------
\section{Exploration of RAG Applications}
\label{app:rag}

To further validate the practical utility of our framework, we extend
our evaluation to Retrieval-Augmented Generation (RAG) scenarios.
Following the experimental setup of LamRA~\citep{liu2024lamra}, we
evaluate our method on three Knowledge-based Visual Question Answering
(KVQA) benchmarks.  Specifically, we train the retrieval and VQA tasks
simultaneously, allowing the model to align the agentic visual
reasoning process with downstream generation needs.  As detailed in
Table~\ref{table_kvqa}, V-Retrver achieves superior performance in
both retrieval precision and VQA accuracy, demonstrating that our
Multimodal Interleaved Evidence Reasoning significantly enhances MLLM
capabilities in RAG settings.

% Table 13 — double-column
\begin{table*}[htbp]
\centering
\small
\caption{\textbf{Comparison of RAG capabilities on KVQA tasks.}}
\label{table_kvqa}
\vspace{4pt}
\setlength{\tabcolsep}{10pt}
\begin{tabular}{lccc}
\toprule
Method
  & \makecell{OKVQA \\ \cite{marino2019ok}}
  & \makecell{Infoseek \\ \cite{chen2023can}}
  & \makecell{E-VQA \\ \cite{mensink2023encyclopedic}} \\
\midrule
\multicolumn{4}{c}{\textit{Retrieval (PR@5)}} \\
\midrule
PreFLMR~\cite{lin2024preflmr}          & 70.9 & 62.1 & 73.7 \\
LamRA-7B~\citep{liu2024lamra}          & 89.0 & 73.4 & 75.0 \\
\rowcolor{gray!20}
V-Retrver-7B                           & \textbf{90.9} & \textbf{78.3} & \textbf{78.1} \\
\midrule
\multicolumn{4}{c}{\textit{VQA (ACC)}} \\
\midrule
RA-VQAv2 w/ PreFLMR~\cite{lin2023fine} & 61.9 & 32.1 & 54.5 \\
LamRA-7B~\citep{liu2024lamra}          & 64.3 & 28.8 & 56.2 \\
\rowcolor{gray!20}
V-Retrver-7B                           & \textbf{65.7} & \textbf{31.9} & \textbf{58.0} \\
\bottomrule
\end{tabular}
\end{table*}

% ==============================================================
% Visual Toolset Extensibility: OCR Tool
% ==============================================================
\section{Visual Toolset Extensibility}
\label{app:ocr_tool}
The current V-Retrver toolset (SELECT-IMAGE and ZOOM-IN) was intentionally kept minimal to validate the MIER framework and ensure stable EAPO policy optimization. To demonstrate the extensibility of our framework to richer perceptual operations, we integrate an additional OCR tool (\textsc{Extract-Text}) into V-Retrver-7B and evaluate under identical settings.
\textsc{Extract-Text} allows the agent to invoke optical character recognition on a specified region of a candidate image, enabling it to acquire exact textual evidence (e.g., labels, captions, product names) that standard visual encodings may fail to capture reliably.
As shown in Table~\ref{tab:ocr_tool}, adding \textsc{Extract-Text} yields a consistent $+0.2\%$ average performance gain. The improvement is most pronounced on the knowledge-intensive OVEN dataset ($+0.4\%$ on $(q^i,q^t){\to}c^t$), where exact text matching in candidate images provides discriminative evidence that visual features alone cannot supply.
\begin{table}[h]
\centering
\small
\caption{Effect of adding an OCR tool (\textsc{Extract-Text}) to V-Retrver-7B under identical training and evaluation settings.}
\label{tab:ocr_tool}
\setlength{\tabcolsep}{4pt}
\resizebox{\columnwidth}{!}{
\begin{tabular}{lccccc}
\toprule
 & $q^t{\to}c^i$ & $q^i{\to}c^t$ & $(q^i,q^t){\to}c^i$ & $(q^i,q^t){\to}c^t$ & \\
\cmidrule(lr){2-2}\cmidrule(lr){3-3}\cmidrule(lr){4-4}\cmidrule(lr){5-5}
Variants & COCO R@5 & F200K R@10 & CIRR R@5 & OVEN R@5 & Avg. \\
\cmidrule(lr){2-2}\cmidrule(lr){3-3}\cmidrule(lr){4-4}\cmidrule(lr){5-5}
\midrule
V-Retrver-7B                    & 87.5 & 37.8 & 73.5 & 69.8 & 67.2 \\
V-Retrver-7B $+$ \textsc{Extract-Text} & \textbf{87.8} & \textbf{37.9} & \textbf{73.7} & \textbf{70.2} & \textbf{67.4} \\
\bottomrule
\end{tabular}
}
\end{table}
These results confirm that the MIER framework and EAPO training pipeline generalize naturally to new tool types: the agent successfully learns \emph{when} to invoke \textsc{Extract-Text} (i.e., when candidate images contain informative text) without requiring task-specific re-engineering of the reward or trajectory format. Expanding the toolset further (for example with object detection or spatial relationship queries) is a promising direction for future work.

% ----------------------------------------------------------------
% SECTION 5 — Implementation Details  (Table 12, single-column)
% ----------------------------------------------------------------
\section{Implementation Details}
\label{app:impl}

\paragraph{Model Initialization.}
V-Retrver is initialized from
Qwen2.5-VL-7B-Instruct~\citep{bai2025qwen2}.  Throughout all
training stages, the vision encoder is frozen while the language model
is fine-tuned.  The embedding model $\phi$ is constructed following
the same method as LamRA~\citep{liu2024lamra}.

\paragraph{Stage~I: Supervised Fine-Tuning.}
We synthesize multimodal Chain-of-Thought data using
Qwen2.5-VL-72B-Instruct, applying rule-based filtering to remove
logically inconsistent or malformed samples.  Training is conducted
using the LLaMA-Factory~\citep{zheng2024llamafactory} framework on
8~A800 GPUs, with a batch size of 64, a learning rate of
$1\times10^{-5}$, and 2 epochs.

\paragraph{Stage~II: Rejection Sampling Fine-Tuning.}
For each training instance, multiple reasoning trajectories are
sampled, and only those satisfying both format validity and correct
retrieval ranking are retained.  Training uses the same framework and
hardware as Stage~I.

\paragraph{Stage~III: Evidence-Aligned Policy Optimization.}
RL training is based on the verl-tool~\citep{jiang2025verltool}
framework, which extends verl~\citep{sheng2024hybridflow} and
vLLM~\citep{kwon2023efficient} to support multimodal
tool-augmented multi-turn training.  The model is trained for 1 epoch
with a learning rate of $1\times10^{-6}$ and 8 rollouts per query.

\paragraph{Reward Hyperparameters.}
Table~\ref{tab:hyperparams} summarizes all reward-related
hyperparameters used in EAPO.

% Table 12 — single-column
\begin{table}[htbp]
\centering
\small
\caption{Hyperparameter configurations for the EAPO training stage.}
\label{tab:hyperparams}
\begin{tabular}{clc}
\toprule
\textbf{Symbol} & \textbf{Description} & \textbf{Value} \\
\midrule
$K_r$    & CNDCG reward cutoff depth          & 5   \\
$m$      & Number of hard negatives for CNDCG & 5   \\
$\alpha$ & Format reward weight (Eq.~1)       & 0.2 \\
$\beta$  & Ranking reward weight (Eq.~1)      & 0.8 \\
$\eta$   & Tool success incentive (Eq.~6)     & 0.2 \\
$\rho$   & Tool excess penalty (Eq.~6)        & 0.1 \\
$\tau$   & Tool tolerance threshold (Eq.~6)   & 1   \\
$\lambda$& KL penalty coefficient (Eq.~8)     & 0   \\
\bottomrule
\end{tabular}
\end{table}

% ----------------------------------------------------------------
% SECTION 6 — Prompt Template
% ----------------------------------------------------------------
\section{Prompt Template}
\label{app:prompt}

\subsection{System Prompt}
\label{promot}
\begin{figure*}[p]
    \centering
    \includegraphics[width=\linewidth]{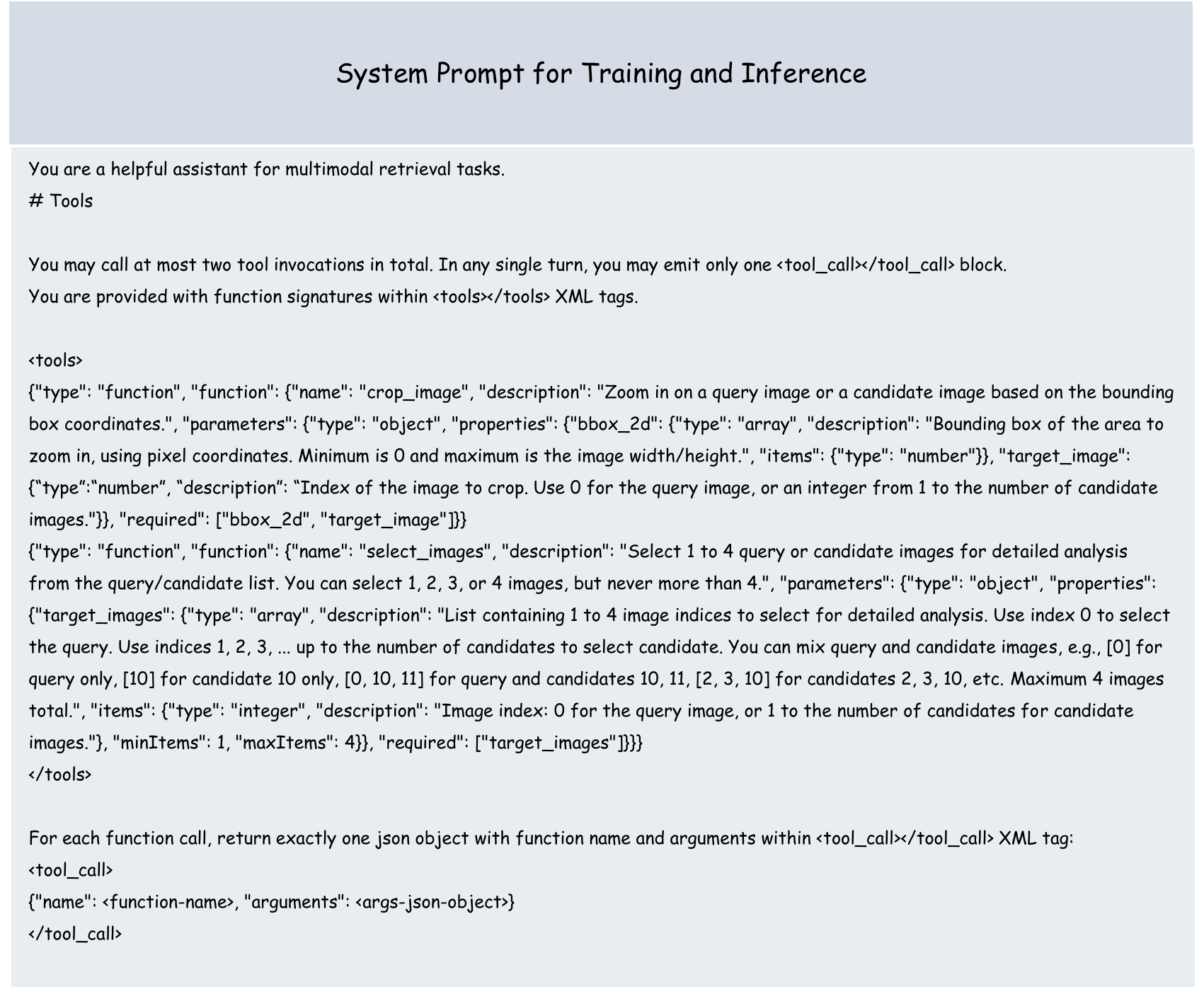}
    \caption{System prompt template for training and inference.}
    \label{fig:prompt_system}
\end{figure*}

Fig.~\ref{fig:prompt_system} illustrates the system prompt used for
both training and inference.

\subsection{User Prompt}
\begin{figure*}[p]
    \centering
    \includegraphics[width=\linewidth]{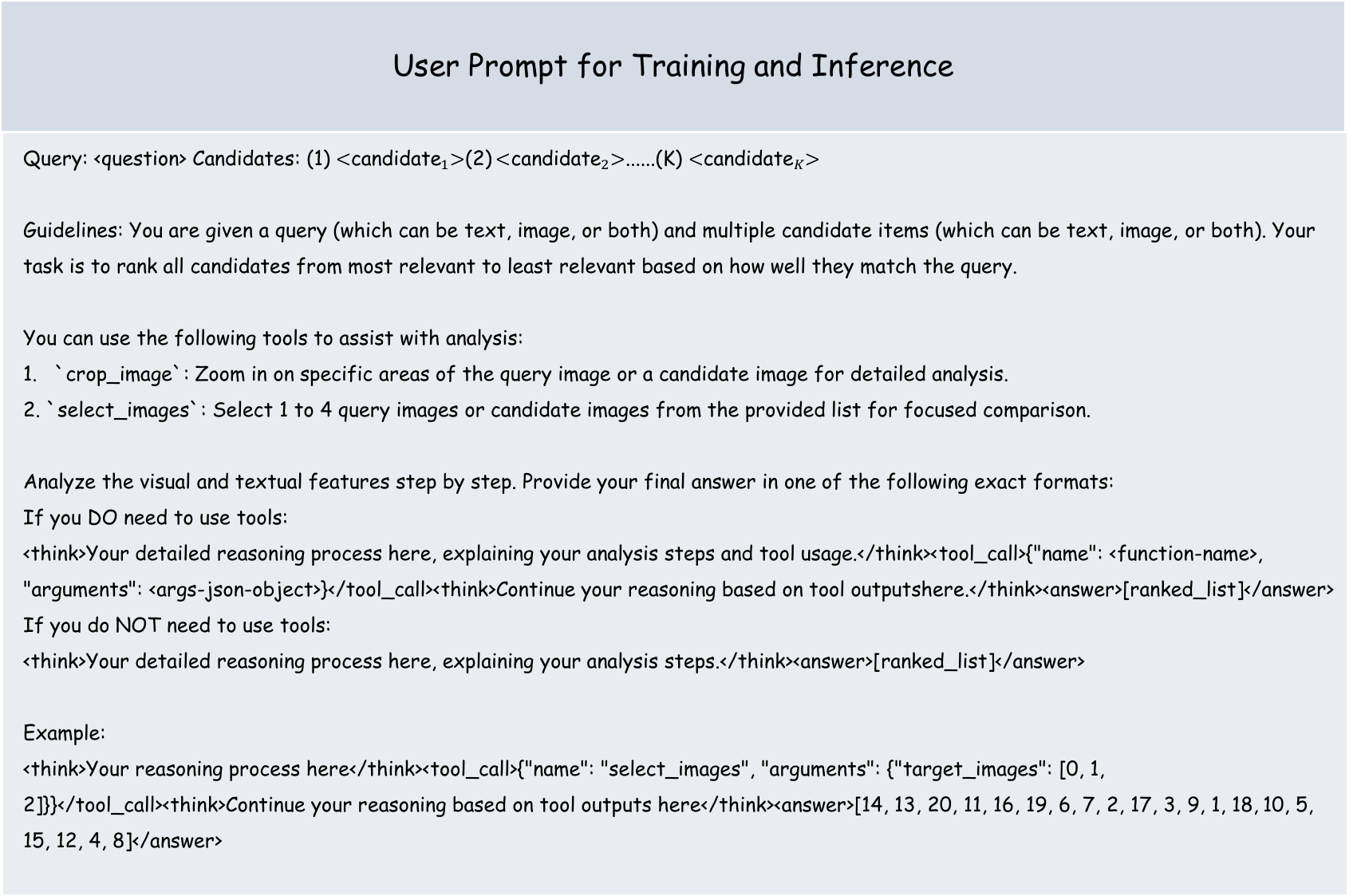}
    \caption{User prompt template for training and inference.}
    \label{fig:prompt_user}
\end{figure*}

Fig.~\ref{fig:prompt_user} illustrates the user prompt used for both
training and inference.

\subsection{Annotation Prompt}
\begin{figure*}[p]
    \centering
    \includegraphics[width=\linewidth]{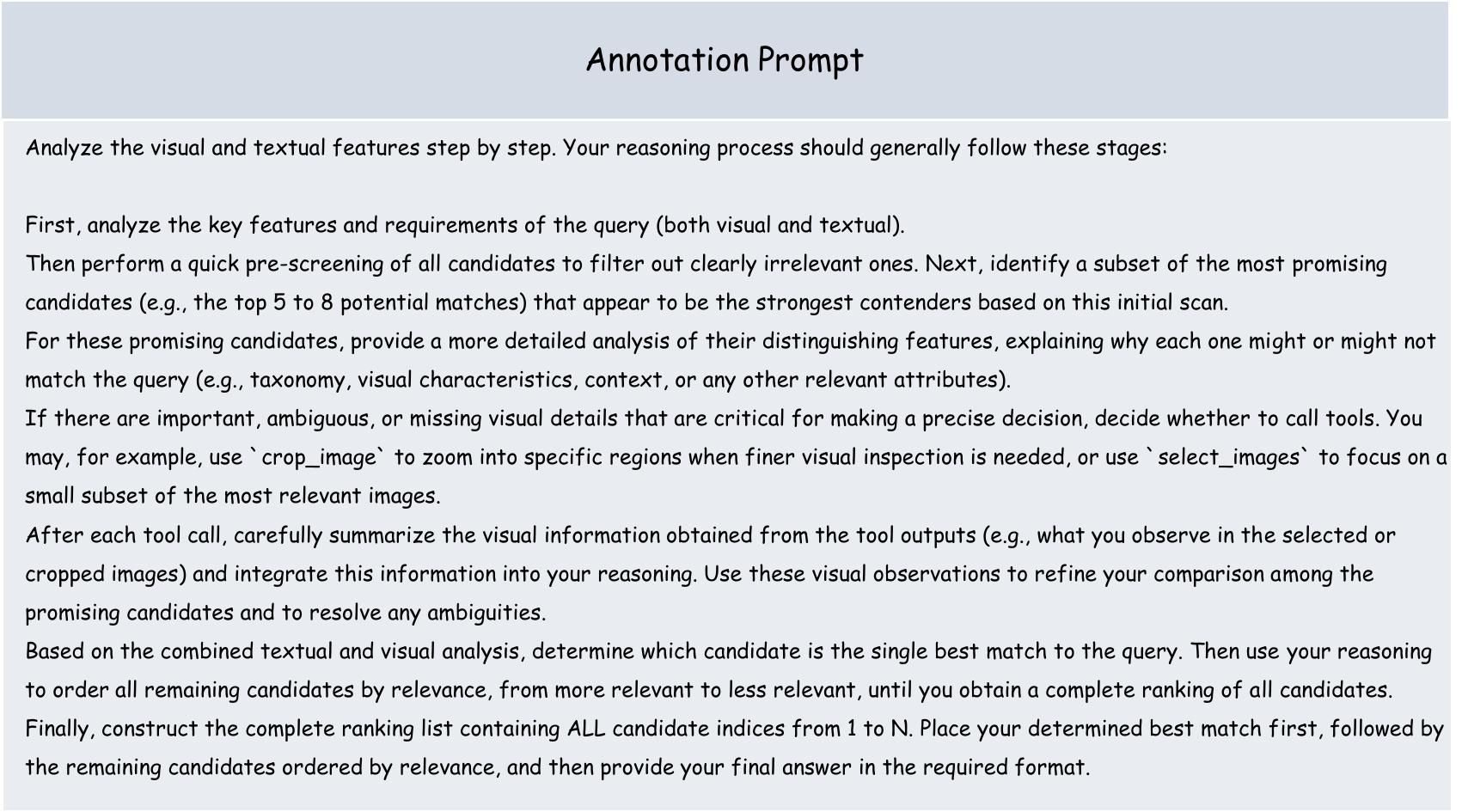}
    \caption{Annotation prompt template.}
    \label{fig:prompt_annotation}
\end{figure*}

Fig.~\ref{fig:prompt_annotation} illustrates the annotation prompt.
Specifically, for the CoT annotation process, this prompt is inserted
into the user prompt to guide generation.

% ----------------------------------------------------------------
% SECTION 7 — Formal Definition of Contrastive NDCG Reward
% ----------------------------------------------------------------
\section{Formal Definition of Contrastive NDCG Reward}
\label{app:cndcg}

\paragraph{Hard Negative Set.}
Given the candidate set $\mathcal{C}$, the hard negative set
$\mathcal{H}$ consists of the top-$m$ non-correct candidates ranked by
embedding similarity to the query:
\begin{equation}
  \mathcal{H}
  = \underset{j \neq j^*,\; |\mathcal{H}|=m}{\mathrm{top}\text{-}m}\;
    \mathrm{sim}\!\left(\phi(c_j),\;\phi(q)\right).
  \label{eq:hard_neg}
\end{equation}
The remaining candidates form the easy negative set
$\mathcal{E} = \mathcal{C} \setminus (\mathcal{H} \cup \{j^*\})$.

The key distinction from vanilla NDCG lies in the contrastive
relevance assignment.  This contrastive property provides targeted
penalization precisely for the failure mode most detrimental to
retrieval quality: a visually similar hard negative outranking the
correct candidate.

% ----------------------------------------------------------------
% SECTION 8 — Dataset Details  (Table 9 single-col, Table 10 double)
% ----------------------------------------------------------------
\section{Dataset Details}
\label{app:datasets}

We present the statistics of all evaluation benchmarks in
Table~\ref{tab:benchmark_summary} and the full breakdown of the M-BEIR
training and evaluation splits in Table~\ref{tab:mbeir_detail}.

% ----------------------------------------------------------------
% Tables 11, 12, 13 — three tables stacked top-to-bottom on one page
% Requires: \usepackage{caption,booktabs,multirow,colortbl,makecell}
% ----------------------------------------------------------------

\begin{table*}[p]
\centering
\small

% ============================================================
% Table 11
% ============================================================
\captionof{table}{Summary of evaluation benchmarks, categorized into
  Supervised and Zero-shot settings. \#~Queries is the number of test
  queries; \#~Candidates is the number of test candidates per query.}
\label{tab:benchmark_summary}
\setlength{\tabcolsep}{10pt}
\begin{tabular}{lrrl}
  \toprule
  \textbf{Benchmark} & \textbf{\#~Queries} & \textbf{\#~Candidates}
                     & \textbf{Setting} \\
  \midrule
  M-BEIR~\citep{wei2024uniir}                          & 190K & 5.6M  & Supervised \\
  \midrule
  CIRCO~\citep{baldrati2023zero}                        & 800  & 120K  & Zero-shot \\
  GeneCIS~\citep{vaze2023genecis}                       & 8K   & 10--15 & Zero-shot \\
  Visual Storytelling~\citep{huang2016visual}           & 5K   & 8K    & Zero-shot \\
  Visual Dialog~\citep{das2017visual}                   & 2K   & 2K    & Zero-shot \\
  Multi-round FashionIQ~\citep{yuan2021conversational}  & 2.4K & 6.2K  & Zero-shot \\
  \bottomrule
\end{tabular}

\bigskip

% ============================================================
% Table 12
% ============================================================
\captionof{table}{Full breakdown of the M-BEIR benchmark by task, dataset, and split.}
\label{tab:mbeir_detail}
\setlength{\tabcolsep}{7pt}
\begin{tabular}{lllrrrr}
  \toprule
  \textbf{Task} & \textbf{Dataset} & \textbf{Domain}
    & \textbf{\#~Train} & \textbf{\#~Dev}
    & \textbf{\#~Test} & \textbf{\#~Pool} \\
  \midrule
  \multirow{3}{*}{$q^t \!\to\! c^i$}
    & VisualNews  & News    & 99K   & 20K   & 20K   & 542K \\
    & MSCOCO      & Misc.   & 100K  & 24.8K & 24.8K & 5K   \\
    & Fashion200K & Fashion & 15K   & 1.7K  & 1.7K  & 201K \\
  \midrule
  $q^t \!\to\! c^t$
    & WebQA       & Wiki    & 16K   & 1.7K  & 2.4K  & 544K \\
  \midrule
  \multirow{2}{*}{$q^t \!\to\! (c^i,c^t)$}
    & EDIS        & News    & 26K   & 3.2K  & 3.2K  & 1M   \\
    & WebQA       & Wiki    & 17K   & 1.7K  & 2.5K  & 403K \\
  \midrule
  \multirow{3}{*}{$q^i \!\to\! c^t$}
    & VisualNews  & News    & 100K  & 20K   & 20K   & 537K \\
    & MSCOCO      & Misc.   & 113K  & 5K    & 5K    & 25K  \\
    & Fashion200K & Fashion & 15K   & 4.8K  & 4.8K  & 61K  \\
  \midrule
  $q^i \!\to\! c^i$
    & NIGHTS      & Misc.   & 16K   & 2K    & 2K    & 40K  \\
  \midrule
  \multirow{2}{*}{$(q^i,q^t) \!\to\! c^t$}
    & OVEN        & Wiki    & 150K  & 50K   & 50K   & 676K \\
    & InfoSeek    & Wiki    & 141K  & 11K   & 11K   & 611K \\
  \midrule
  \multirow{2}{*}{$(q^i,q^t) \!\to\! c^i$}
    & FashionIQ   & Fashion & 16K   & 2K    & 6K    & 74K  \\
    & CIRR        & Misc.   & 26K   & 2K    & 4K    & 21K  \\
  \midrule
  \multirow{2}{*}{$(q^i,q^t) \!\to\! (c^i,c^t)$}
    & OVEN        & Wiki    & 157K  & 14.7K & 14.7K & 335K \\
    & InfoSeek    & Wiki    & 143K  & 17.6K & 17.6K & 481K \\
  \midrule
  \textbf{Total} & 10 datasets & 4 domains
    & \textbf{1.1M} & \textbf{182K} & \textbf{190K} & \textbf{5.6M} \\
  \bottomrule
\end{tabular}

\bigskip

\caption{Summary of unseen datasets used in zero-shot evaluation (Setting~II).}
\label{tab:unseen_datasets}
\setlength{\tabcolsep}{4pt}
\begin{tabular}{llll}
\toprule
\textbf{Dataset} & \textbf{Image Source} & \textbf{Task} & \textbf{Query Format} \\
\midrule
CIRCO         & MSCOCO (unlabeled) & $(q^i,q^t)\!\to\!c^i$           & \texttt{<image><rel.\ caption>} \\
GeneCIS       & MSCOCO             & $(q^i,q^t)\!\to\!c^i$           & \texttt{<image><rel.\ caption>} \\
Visual Dialog & MSCOCO             & $q^\text{dialog}\!\to\!c^i$     & \texttt{<Q1><A1>...<Qj>} \\
Vis.\ Story.  & Flickr             & $(q^i\!\oplus\!q^t)\!\to\!c^i$ & \texttt{<text><image>...<text>} \\
MT-FashionIQ  & FashionIQ          & $(q^i\!\oplus\!q^t)\!\to\!c^i$ & \texttt{<image><cap>...<cap>} \\
\bottomrule
\end{tabular}

\end{table*}

\section{Detailed Experiment Settings}
\label{app:settings}
\label{app:baselines}

We establish three distinct experiment settings to comprehensively
evaluate V-Retrver.

\paragraph{Setting~I: Supervised Evaluation on M-BEIR.}
To validate the versatility of our method, we train V-Retrver on all
8~tasks in M-BEIR and evaluate on the corresponding test sets.
Baselines include:
(1)~foundational VLMs: Qwen2.5-VL-3B/7B, CLIP-L, SigLIP, BLIP,
BLIP2;
(2)~fine-tuned universal retrievers: UniIR-BLIP$_\text{FF}$,
UniIR-CLIP$_\text{SF}$;
(3)~reasoning-enhanced and universal retrieval models:
Vision-R1~\citep{huang2025vision},
VLM-R1~\citep{shen2025vlm},
MM-Embed~\citep{lin2024mm},
LamRA~\citep{liu2024lamra},
U-MARVEL~\citep{li2025u},
Retrv-R1~\citep{zhu2025retrv};
and (4)~agentic tool-use models: DeepEyes.

\paragraph{Setting~II: Zero-shot Evaluation on Unseen Datasets.}
To evaluate generalization on previously unseen retrieval datasets, we
perform zero-shot experiments on 5~datasets not encountered during
training: CIRCO, GeneCIS, Visual Storytelling, Visual Dialog, and
Multi-round FashionIQ.  Baselines include:
E5-V~\citep{jiang2024e5},
MagicLens~\citep{zhang2024magiclens},
MM-Embed~\citep{lin2024mm},
LamRA~\citep{liu2024lamra}.

\paragraph{Setting~III: Held-out Task Evaluation.}
To investigate generalization to unseen retrieval tasks, we
intentionally exclude three task types during training:
image-to-image retrieval ($q^i \!\to\! c^i$),
text-image-to-text retrieval ($(q^i,q^t) \!\to\! c^t$),
and text-image-to-text-image retrieval
($(q^i,q^t) \!\to\! (c^i,c^t)$).
Training is conducted on the remaining five tasks, and evaluation is
performed on the three held-out tasks.  In this zero-shot setting, we
compare against Qwen2.5-VL-7B, Vision-R1, and LamRA$^*$ (also trained
on the five non-held-out tasks for a fair comparison).

The unseen datasets used in Setting~II are summarized in
Table~\ref{tab:unseen_datasets}.

% ----------------------------------------------------------------
% SECTION 10 — Algorithms
% ----------------------------------------------------------------
\section{Algorithms and Detailed Analysis}
\label{app:algorithms}

We present the formal algorithms for the inference and training
processes of V-Retrver.  The inference process, formulated as a
coarse-to-fine pipeline with sliding window agentic reasoning, is
detailed in Algorithm~\ref{alg:inference}.  The three-stage curriculum
learning strategy, designed to progressively align the model with
evidence-driven retrieval objectives, is presented in
Algorithm~\ref{alg:training}.

\begin{figure*}[htbp]
\begin{minipage}[t]{0.48\textwidth}
\begin{algorithm}[H]
\caption{V-Retrver Inference Pipeline}
\label{alg:inference}
\begin{algorithmic}[1]
    \STATE \textbf{Input:} Query $q$, Candidate Pool
        $\Omega{=}\{c_n\}_{n=1}^N$, Embedding Model $\Phi$,
        Reasoning Agent $\pi_\theta$, Top-$K$ size $K$,
        Window size $W$, Stride $S$
    \STATE \textbf{Output:} Ranked Candidate List $\hat{L}$
    \STATE \COMMENT{Stage 1: Coarse Retrieval}
    \STATE Compute $s_n = \cos(\Phi(q), \Phi(c_n))$ for all
        $c_n \in \Omega$
    \STATE $\mathcal{C}_{\text{top}} \leftarrow
        \text{Top-K}(\Omega, \{s_n\})$
    \STATE \COMMENT{Stage 2: Agentic Reranking}
    \STATE Initialize $\mathcal{L}_{\text{global}} \leftarrow \emptyset$
    \STATE Split $\mathcal{C}_{\text{top}}$ into windows
        $\{w_1,\dots,w_m\}$
    \FOR{each window $w_j$}
        \STATE $H_0 \leftarrow (q,\, w_j,\, \text{Instruction})$,
            \quad $t \leftarrow 0$
        \WHILE{True}
            \STATE Generate output: $o_t \sim \pi_\theta(H_t)$
            \IF{$o_t$ contains \texttt{<tool\_call>}}
                \STATE Parse $a_t$ from $o_t$
                \STATE $v_{\text{obs}} \leftarrow
                    f_{\text{tool}}(a_t, w_j)$
                \STATE $H_{t+1} \leftarrow H_t \oplus o_t
                    \oplus v_{\text{obs}}$
            \ELSIF{$o_t$ contains \texttt{<answer>}}
                \STATE Parse $\hat{r}_j$ from $o_t$
                \STATE Update $\mathcal{L}_{\text{global}}$ with
                    $\hat{r}_j$;\quad \textbf{break}
            \ENDIF
            \STATE $t \leftarrow t + 1$
        \ENDWHILE
    \ENDFOR
    \STATE $\hat{L} \leftarrow
        \text{AggregateRanks}(\mathcal{L}_{\text{global}})$
\end{algorithmic}
\end{algorithm}
\end{minipage}
\hfill
\begin{minipage}[t]{0.48\textwidth}
\begin{algorithm}[H]
\caption{Curriculum-Based Agentic Training}
\label{alg:training}
\begin{algorithmic}[1]
    \STATE \textbf{Input:} Pretrained MLLM $\theta_{\text{init}}$,
        Dataset $\mathcal{D}$, Synth Model $M_{\text{syn}}$
    \STATE \textbf{Output:} Optimized Policy $\pi_{\theta^*}$
    \STATE \COMMENT{Stage 1: Reasoning Activation (SFT)}
    \STATE $\mathcal{D}_{\text{sft}} \leftarrow
        \{(q,c,\tau_{\text{cot}})\}$ via $M_{\text{syn}}$ on
        $\mathcal{D}$
    \STATE Filter $\mathcal{D}_{\text{sft}}$ for format compliance
    \STATE $\theta_{\text{sft}} \leftarrow
        \text{Minimize}\;
        \mathcal{L}_{\text{SFT}}(\theta_{\text{init}},
        \mathcal{D}_{\text{sft}})$
    \STATE \COMMENT{Stage 2: Reliability Refinement (RSFT)}
    \STATE Initialize $\mathcal{D}_{\text{rsft}} \leftarrow \emptyset$
    \FOR{each $(q, c) \in \mathcal{D}$}
        \STATE Sample $\{\tau_1,\dots,\tau_k\} \sim
            \pi_{\theta_{\text{sft}}}(q,c)$
        \IF{$\text{IsFormatValid}(\tau_i)
            \land \text{IsRankCorrect}(\tau_i)$}
            \STATE Add $\tau_i$ to $\mathcal{D}_{\text{rsft}}$
        \ENDIF
    \ENDFOR
    \STATE $\theta_{\text{rsft}} \leftarrow
        \text{Minimize}\;
        \mathcal{L}_{\text{SFT}}(\theta_{\text{sft}},
        \mathcal{D}_{\text{rsft}})$
    \STATE \COMMENT{Stage 3: EAPO}
    \STATE $\theta \leftarrow \theta_{\text{rsft}}$,\quad
        $\pi_{\text{ref}} \leftarrow \theta_{\text{rsft}}$
    \WHILE{not converged}
        \STATE Sample $B_q \sim \mathcal{D}$
        \FOR{each $q \in B_q$}
            \STATE Sample $\{o_1,\dots,o_G\} \sim \pi_{\theta}(q)$
            \STATE $R(o_i) = \alpha r_{\text{fmt}}
                + \beta r_{\text{rank}} + r_{\text{tool}}$
        \ENDFOR
        \STATE Compute $A_i$ via group normalization
        \STATE Update $\theta \leftarrow
            \text{Optimize}\;
            \mathcal{J}_{\text{EAPO}}(\theta)$
    \ENDWHILE
    \STATE \textbf{return} $\pi_{\theta}$
\end{algorithmic}
\end{algorithm}
\end{minipage}
\end{figure*}

% ----------------------------------------------------------------
% SECTION 11 — Qualitative Examples
% ----------------------------------------------------------------
\section{Qualitative Examples}
\label{app:qualitative}

\begin{figure*}[p]
    \centering
    \includegraphics[width=\linewidth]{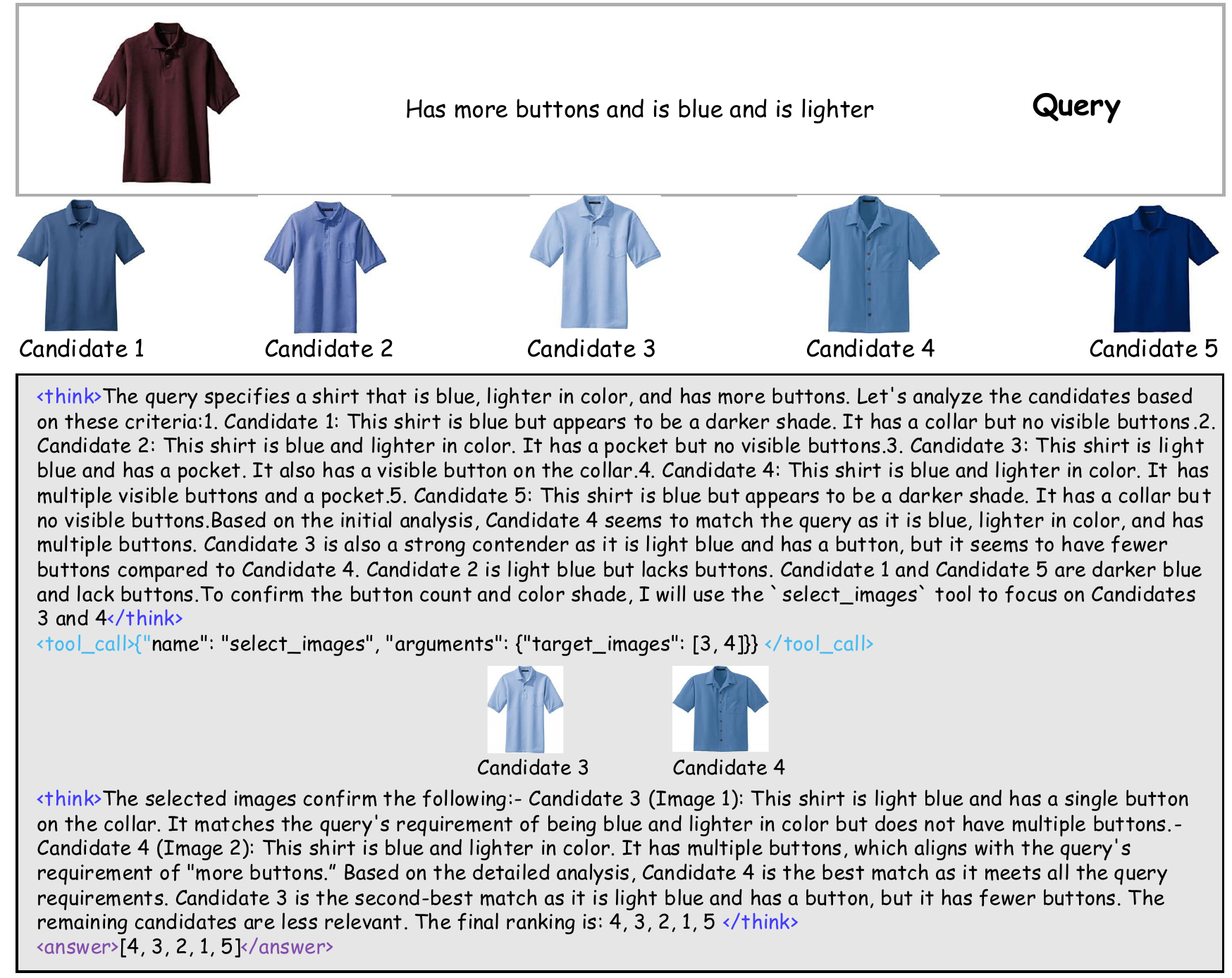}
    \caption{A qualitative example of the retrieval result generated by
      V-Retrver.}
    \label{fig:example1}
\end{figure*}

\begin{figure*}[p]
    \centering
    \includegraphics[width=\linewidth]{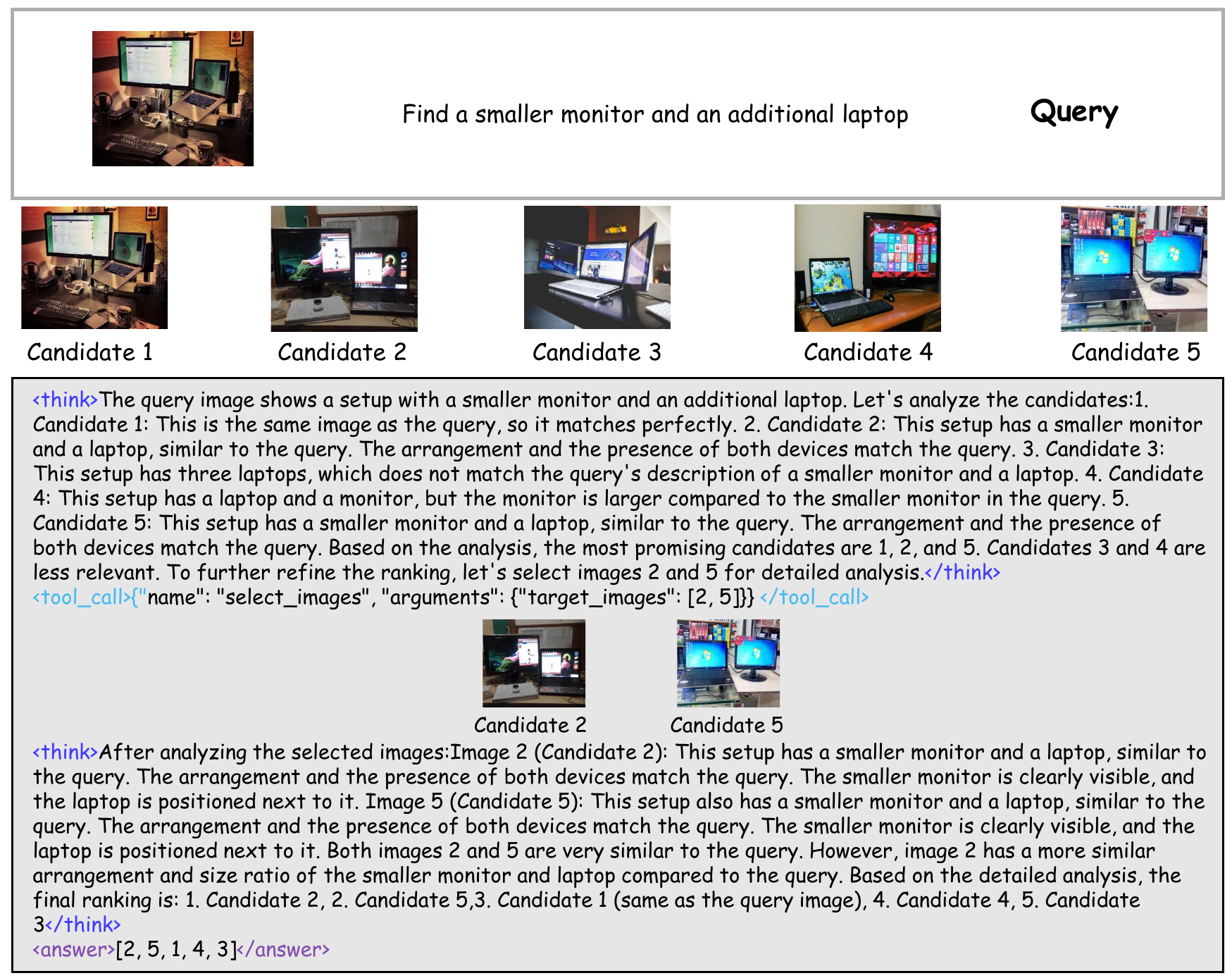}
    \caption{A qualitative example of the retrieval result generated by
      V-Retrver.}
    \label{fig:example2}
\end{figure*}

\begin{figure*}[p]
    \centering
    \includegraphics[width=\linewidth]{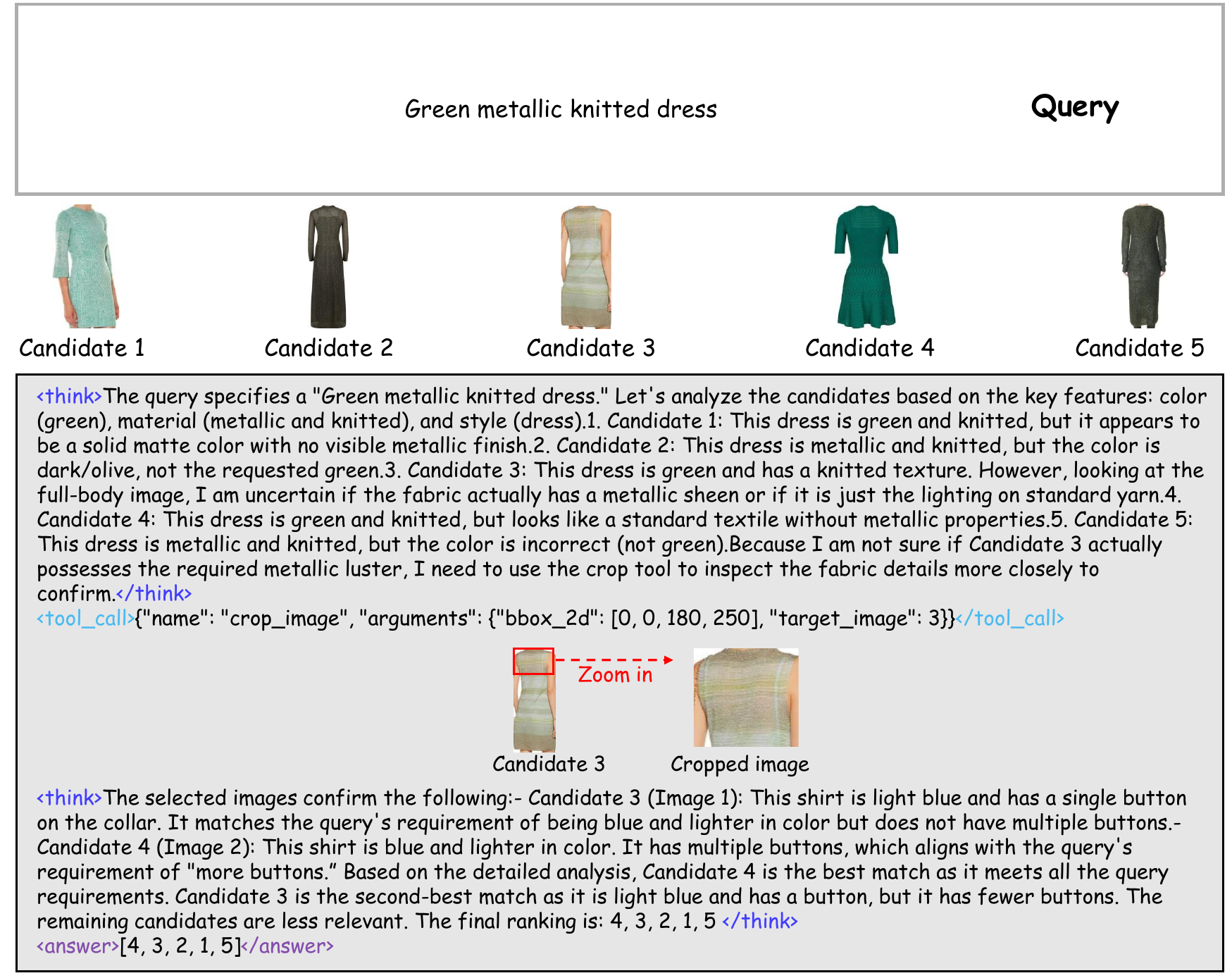}
    \caption{A qualitative example of the retrieval result generated by
      V-Retrver.}
    \label{fig:example3}
\end{figure*}

\begin{figure*}[p]
    \centering
    \includegraphics[width=\linewidth]{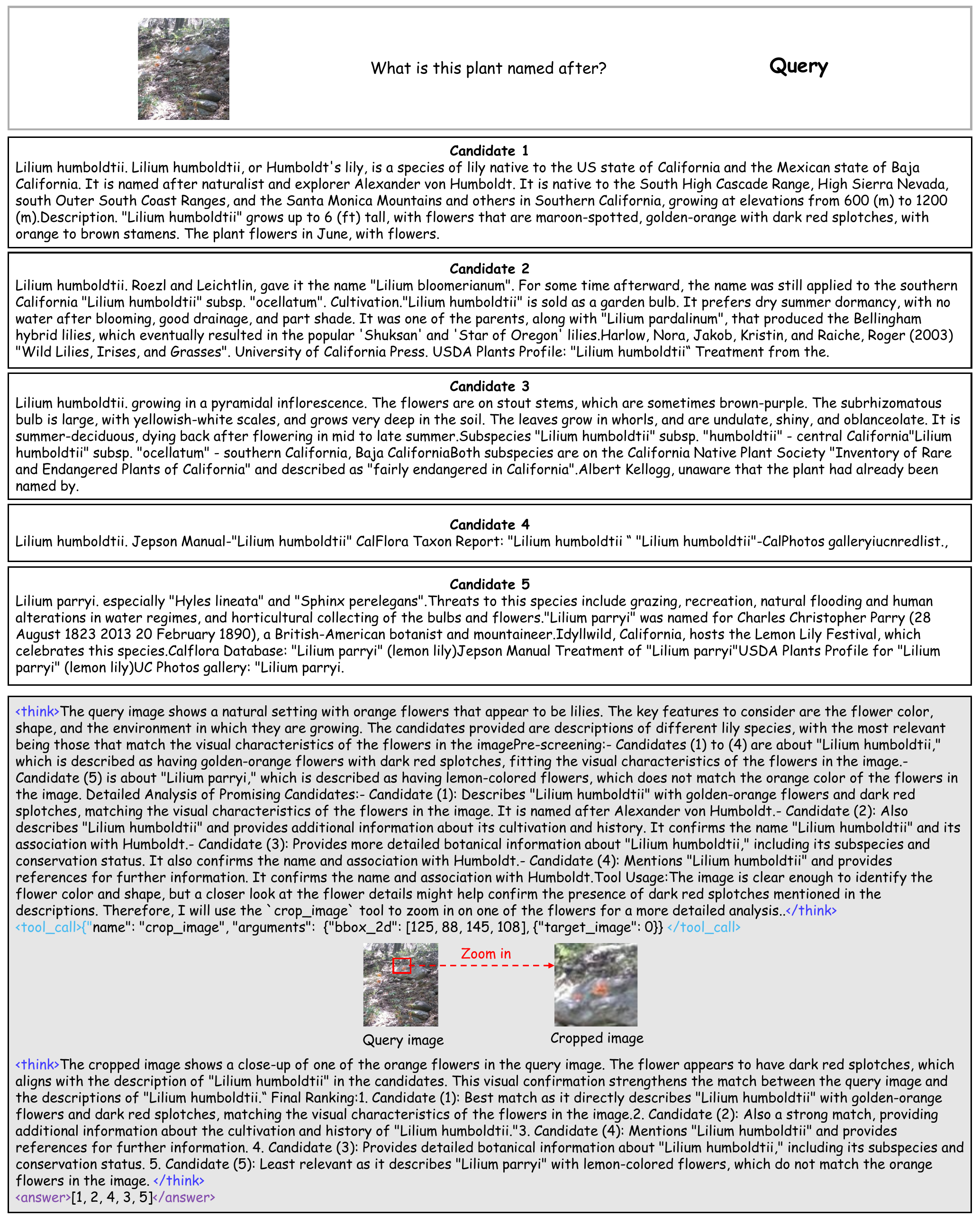}
    \caption{A qualitative example of the retrieval result generated by
      V-Retrver.}
    \label{fig:example4}
\end{figure*}

\begin{figure*}[p]
    \centering
    \includegraphics[width=\linewidth]{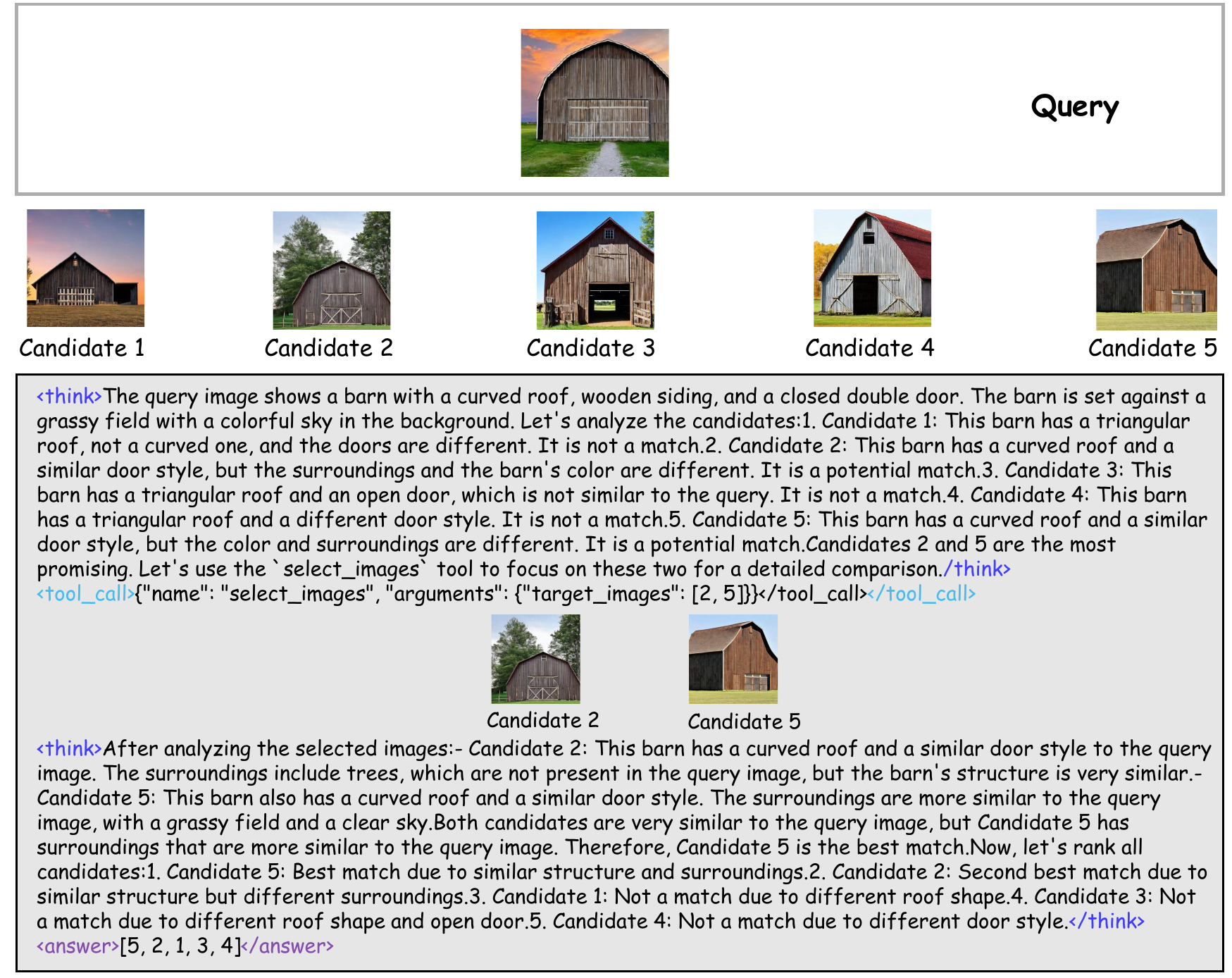}
    \caption{A qualitative example of the retrieval result generated by
      V-Retrver.}
    \label{fig:example5}
\end{figure*}

To provide an intuitive illustration of our approach and further
demonstrate the effectiveness of V-Retrver, we present qualitative
results in Figs.~\ref{fig:example1} through \ref{fig:example5}.  These
examples illustrate how V-Retrver performs accurate retrieval through
fine-grained and structured reasoning, highlighting the strong
effectiveness of the proposed method.

% \section{Example Appendix}
% \label{sec:appendix}

% This is an appendix.

\end{document}